\documentclass[11pt]{article}

\PassOptionsToPackage{table}{xcolor}
\usepackage{acl}
\onecolumn
\renewcommand{\twocolumn}[1][]{#1}

\usepackage{times}
\usepackage{comment}
\usepackage{caption}
\usepackage{etoolbox}
\AtBeginEnvironment{table}{\captionsetup{width=0.48\textwidth}}
\AtBeginEnvironment{table*}{\captionsetup{width=\textwidth}}
\usepackage{float}
\usepackage{latexsym}
\usepackage[T1]{fontenc}
\usepackage[utf8]{inputenc}
\usepackage{microtype}
\usepackage{inconsolata}
\usepackage{graphicx}
\graphicspath{{../outputs/}{../figures/}{figures/}}
\usepackage{amsmath,amssymb,amsfonts}
\usepackage{mathtools}
\usepackage{booktabs}
\usepackage{multirow}
\usepackage{xcolor}
\usepackage[section]{placeins}

\usepackage{colortbl}
\usepackage{tikz}
\usetikzlibrary{shapes.geometric,arrows.meta,positioning,fit,backgrounds,calc}
\usepackage{subcaption}
\usepackage{enumitem}
\usepackage{url}
\usepackage{amsthm}
\usepackage{arydshln} 
\usepackage{todonotes}
\newtheorem{theorem}{Theorem}

\definecolor{lightgray}{gray}{0.93}
\definecolor{lightgreen}{RGB}{220,255,220}
\definecolor{lightred}{RGB}{255,230,230}



\usepackage{lineno}
\newcommand{\1}{\mathbf{1}}

\definecolor{redflag}{HTML}{D32F2F}
\definecolor{purpleflag}{HTML}{7B1FA2}
\definecolor{purplebg}{HTML}{BBDEFB}
\definecolor{blueflag}{HTML}{1565C0}
\DeclareMathOperator*{\argmin}{arg\,min}

\newenvironment{keywords}{\paragraph*{Keywords}}{}

\title{\textsc{CARE}: A Conformal Safety Layer for Medical Summarization}
\author{
\mdseries
Suhana Bedi$^{1,*}$ \quad Bridget Lin$^{1,*}$ \quad Anson Y. Zhou$^{1}$ \quad Chloe O. Stanwyck$^{1,2}$ \\
Jenelle A. Jindal \quad Sanmi Koyejo$^{3,\dagger}$ \quad David Stutz$^{4,\dagger}$ \quad Nigam H. Shah$^{5,\dagger}$ \\
\normalsize
\\
$^{1}$Department of Biomedical Data Science, Stanford University \\
$^{2}$Department of Anesthesiology, Perioperative and Pain Medicine, Stanford University \\
$^{3}$Department of Computer Science, Stanford University \\
$^{4}$Google DeepMind \quad
$^{5}$Department of Medicine, Stanford University \\ [0.5em]
{\normalfont\footnotesize
\textsuperscript{*}Equal contribution. \quad
\textsuperscript{\textdagger}Equal supervision.}
}

\begin{document}
\maketitle

\begin{abstract}
Large language models (LLMs) are increasingly used for medical summarization but can omit medically important information and introduce unsupported claims. Existing error-detection methods produce heuristic or uncalibrated scores, providing no formal control over missed errors and no principled way to trade off safety against clinician review burden. We introduce Conformal Assessment for Risk Evaluation (\textsc{CARE}), a post-hoc risk-control layer that adds calibrated omission and hallucination annotations without retraining the summarizer. \textsc{CARE} uses scalar conformal risk control (CRC) to bound the probability that a document contains an unflagged hallucinated sentence, and Learn-Then-Test fixed-sequence testing (LTT-FST) to control the expected fraction of true omissions left unsurfaced for review. Omission surfacing depends jointly on sentence importance and summary coverage, a two-dimensional threshold-selection problem; we show that calibrating one gate and then imposing an uncalibrated second gate can invalidate risk control. Across five medical summarization tasks, mean held-out violation over 100 random calibration/test resplits remains at or below the target for both controllers. At $\alpha=0.15$, LTT-FST achieves the lowest review workload on three of five datasets among omission baselines held to the same confidence guarantee. Review workload varies by task and is highest for long, high-compression documents. On 46 clinician-annotated documents, the calibrated hallucination controller flags 88\% of clinician-identified hallucinated sentences. In a preliminary study of 75 clinician reviews, assisted review-time omission coverage increased from 50.4\% to 79.0\% without increasing review time. Overall, \textsc{CARE} provides a principled framework for targeted, risk-controlled review of LLM-generated medical summaries.
\end{abstract}


% ============================================================================
\section{Introduction}
\label{sec:intro}

\begin{figure*}[!t]
\centering
\resizebox{0.92\textwidth}{!}{\includegraphics{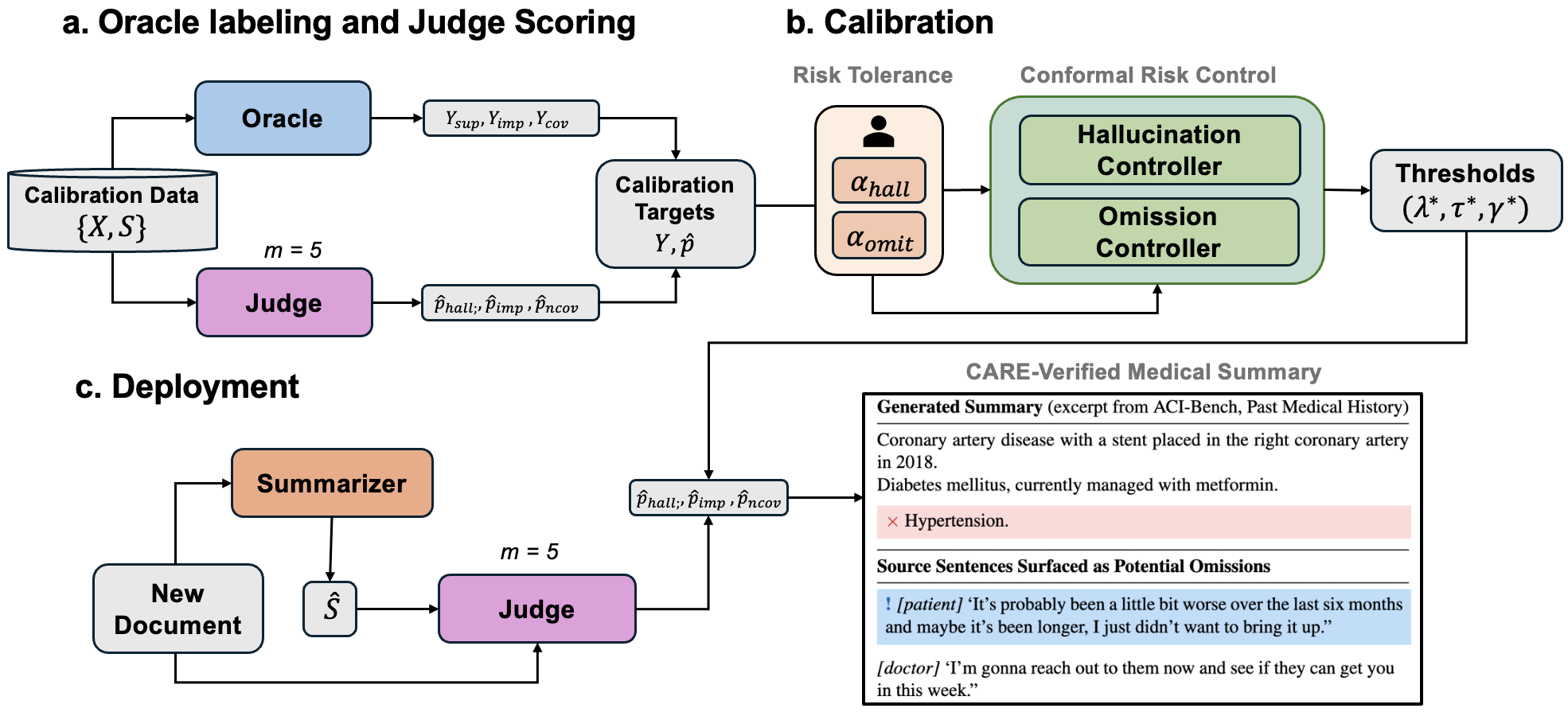}}
\caption{\textbf{CARE pipeline.}
\textsc{CARE} uses scalar conformal risk control (CRC) for hallucination flagging and Learn-Then-Test fixed-sequence testing (LTT-FST) for two-dimensional omission calibration, converting oracle labels and judge scores into risk-controlled sentence-level review annotations.}
\label{fig:schematic}
\end{figure*}

Large language models (LLMs) are rapidly being adopted for medical documentation and summarization. LLM-based systems can draft clinical notes from physician--patient conversations \citep{jama-ambient-scribes-2025}, convert radiology findings into impressions \citep{van2023radadapt}, and generate discharge summaries \citep{lancet-russell-gpt-2024}. Despite their promise, two failure modes remain central:

\begin{enumerate}[nosep,leftmargin=*]
\item \textbf{Omissions}: medically important content absent from the summary. Reported rates range from 3--4\% in controlled settings \citep{npj-clinical-safety-2025} to 47\% in encounter-level evaluations \citep{williams2025}.
\item \textbf{Hallucinations}: content unsupported by the source. Reported rates range from 1--2\% in controlled note generation \citep{npj-clinical-safety-2025} to 42\% of summaries containing at least one hallucinated statement \citep{williams2025}.
\end{enumerate}

Calibration and uncertainty remain inconsistently assessed in healthcare LLM evaluations \citep{bedi2025testing}. Current detectors rely on heuristic scoring or LLM-as-a-judge evaluation \citep{factscore-2023,summac-2022,verifact-2025,oukelmoun-omission-emnlp-2025} and give no principled way to choose thresholds or control missed-error risk: permissive thresholds miss errors, and conservative ones erase the efficiency gains of AI-assisted documentation. Prior conformal methods for text generation intervene at the response level through abstention or rewriting \citep{angelopoulos-crc-2024,quach-clm-2024,wang-etal-2024-conu}, whereas medical summarization benefits from targeted review that preserves the draft \citep{gero2024attributes}.

We present \textsc{CARE} (\textbf{C}onformal \textbf{A}ssessment for \textbf{R}isk \textbf{E}valuation), a post-hoc, model-agnostic risk-control layer that can be recalibrated for different summarizers and domains. Its \textbf{omission controller} surfaces source sentences that are important and likely missing, controlling the expected per-document \emph{fraction} of true omissions left unsurfaced while minimizing the source text clinicians must review. Its \textbf{hallucination controller} flags unsupported summary sentences, controlling the probability that a document contains \emph{any} unflagged hallucinated sentence.

\textbf{Our major contributions are:}
\begin{enumerate}[nosep,leftmargin=*]
\item \textbf{Framework:} A post-hoc risk-control layer that adds sentence-level omission and hallucination annotations to LLM-generated medical summaries without retraining the summarizer.
\item \textbf{Method:} A two-dimensional importance--coverage formulation for omission control, calibrated with a pre-specified Learn-Then-Test fixed-sequence testing (LTT-FST) procedure that preserves finite-sample validity without post-hoc grid optimization.
\item \textbf{Evaluation:} Across five datasets, both controllers keep mean held-out violation at or below the target, and LTT-FST achieves the lowest review workload on three of five datasets among omission baselines held to the same confidence guarantee. We compare against conformal importance selection, document-level conformal abstention, and off-the-shelf faithfulness detectors, and validate calibrated thresholds against clinician labels.
\item \textbf{Workflow evaluation:} In a preliminary study of 75 clinician reviews, \textsc{CARE} increased assisted review-time omission coverage from 50.4\% to 79.0\% with no increase in review time, while clinicians' own detection rose from 50.4\% to 60.3\%.
\end{enumerate}

\section{Related Work}
\label{sec:related}

\subsection{Omission and Hallucination Detection in Medical Summarization.}
Omission detection remains comparatively underdeveloped, with recent work using density-based scoring or structured coverage \citep{oukelmoun-omission-emnlp-2025,zhang-etal-2022}. Hallucination detection includes NLI- and QA-based scorers such as \textsc{SummaC} \citep{summac-2022}, \textsc{FactScore} \citep{factscore-2023}, and \textsc{QAFactEval} \citep{qafacteval}, rubric-guided LLM-as-a-judge evaluators \citep{liu-etal-2023-g, fu2023gptscoreevaluatedesire}, and medical adaptations such as \textsc{VeriFact} \citep{verifact-2025}, which verifies atomic facts against the EHR, and \textsc{MedFactEval} \citep{grolleau2025medfacteval}, which uses an LLM jury. These are evaluation tools: they produce scores or labels rather than calibrated review policies, and they provide no formal control over missed errors. \textsc{CARE} converts such scores into calibrated sentence-level flags.

\subsection{Conformal Prediction for Text Generation.}
Conformal prediction and conformal risk control (CRC) provide distribution-free, finite-sample guarantees on user-specified losses \citep{shafer2007tutorialconformalprediction, angelopoulos-crc-2024}. Most conformal methods for text generation abstain from entire outputs when uncertainty is high \citep{quach-clm-2024, abbasi-yadkori-abstention-2024, wang-etal-2024-conu}, which is poorly suited to long medical summaries: one uncertain claim can discard the whole draft. \citet{mohri-conformal-factuality-2024} instead calibrate a claim-level threshold and remove low-confidence claims. \textsc{CARE}'s hallucination controller uses the same document-level calibration event at the sentence level but flags sentences for review rather than deleting them. \citet{kuwahara2025} conformally calibrate sentence importance so that extractive summaries retain a target fraction of salient content; medical summaries are abstractive \citep{van2023radadapt, wu-etal-2023-knowlab}, and that work does not address hallucinations or review workload. \citet{pang2025conformal} filter hallucinations without labels under weaker batch-level guarantees, \citet{shrestha-kim-clin-conformal-2026} give FDR-style precision guarantees for clinical entity extraction, and \citet{scope-2026} calibrate pairwise LLM judges with abstention. Learn-Then-Test (LTT) treats candidate policies as hypotheses under family-wise error control \citep{angelopoulos-ltt-2025}.

\textsc{CARE}'s methodological contribution is valid joint importance--coverage selection for omissions; the hallucination controller uses established scalar CRC to complete the workflow. We compare directly against the closest prior methods: importance-only conformal selection in the style of \citet{kuwahara2025} (Section~\ref{sec:baselines}), document-level conformal abstention (Section~\ref{sec:sent-vs-doc}), and off-the-shelf faithfulness detectors at default or F1-optimal thresholds (Section~\ref{sec:detectors}).

\section{Method}
\label{sec:method}

\textsc{CARE} constructs oracle labels offline, obtains sentence scores from a cheaper judge, calibrates risk-controlled policies, and applies them at deployment (Figure~\ref{fig:schematic}).

\subsection{Problem Setup}

Given source text $X$, a summarizer $f$ produces an abstractive summary $\hat{S}=f(X)$, with source sentences $U(X)=\{u_1,\dots,u_M\}$ and summary sentences $V(X)=\{v_1,\dots,v_N\}$. We control two document-level risks with separate guarantees rather than a single joint guarantee. \textbf{Hallucination risk} is the probability that a document contains at least one unsupported summary sentence that is not flagged; scalar CRC controls it in expectation. \textbf{Omission risk} is the expected per-document fraction of true omissions not surfaced for review; LTT-FST controls it with confidence at least $1-\delta_{\mathrm{omit}}$ over calibration. Documents with no true omissions contribute zero omission loss, so this is a population-average target, not a guarantee that every document is omission-free. As a common sufficient condition, we assume calibration and deployment documents are i.i.d.\ from the same population. The guarantees are marginal and do not provide document- or subgroup-conditional control; formal statements, and the weaker exchangeability condition sufficient for scalar CRC, are in Appendix~\ref{app:guarantee}.

\subsection{Phase 1: Oracle Labeling}
\label{sec:phase1}

A high-capacity oracle LLM constructs binary labels for calibration. \textbf{Support} $Y_{\text{sup}}(v_i)\in\{0,1\}$ indicates whether summary sentence $v_i$ is supported by $X$ (hallucinations have $Y_{\text{sup}}=0$). \textbf{Importance} $Y_{\text{imp}}(u_j)\in\{0,1\}$ indicates whether source sentence $u_j$ should be retained in a high-quality summary, using the dataset's human-authored reference summary $S$ as an anchor. \textbf{Coverage} $Y_{\text{cov}}(u_j)\in\{0,1\}$ indicates whether $u_j$ is represented in $\hat S$, allowing paraphrase. The true omissions are $O_{\text{true}}(X)=\{u_j: Y_{\text{imp}}(u_j)=1,\ Y_{\text{cov}}(u_j)=0\}$.

We validate oracle labels against two independent clinicians on 46 documents, at the granularity each loss uses (Appendix~\ref{app:oracle-validation}). For the joint omission label, oracle--clinician F1 is 0.80 on MIMIC-BHC and 0.66 on ACI-Bench, versus clinician--clinician F1 of 0.85 and 0.72. Document-level hallucination agreement between oracle and clinicians is comparable to inter-clinician agreement on all three annotated datasets; sentence-level hallucinations are rare, and agreement on them is lower for every pair. Guarantees are therefore stated with respect to the calibration labels. The labeler is a plug-in: a deployment that prefers clinician-defined errors can calibrate directly on clinician labels.

\subsection{Phase 2: Scoring by a Judge Model}
\label{sec:phase2}

At deployment, a cheaper judge provides graded evidence using three-level rubrics (Supported/Partial/Unsupported, Essential/Relevant/Not Relevant, Covered/Partial/Omitted; \citealp{hong2026rulers}). We query $m=5$ stochastic replicates per sentence, map labels, in rubric order, to $\{1,0.5,0\}$, and average, giving support, importance, and coverage scores $\hat p_{\mathrm{sup}}(v_i)$, $\hat p_{\mathrm{imp}}(u_j)$, and $\hat p_{\mathrm{cov}}(u_j)$, with non-coverage $\hat p_{\mathrm{ncov}}=1-\hat p_{\mathrm{cov}}$. Low support indicates likely hallucination; high importance and high non-coverage indicate likely omission.

\subsection{Phase 3: Risk-Controlled Calibration}
\label{sec:phase3}

\paragraph{Hallucination controller.}
Summary sentence $v_i$ is flagged when $\hat{p}_{\text{sup}}(v_i) \le \lambda$. The document-level loss $L_{\text{hall}}(\lambda,X)$ is 1 if at least one hallucinated sentence is not flagged and 0 otherwise. CRC selects the smallest $\lambda^*$ (fewest flags) satisfying $\frac{n}{n+1}\hat{R}_{\text{hall}}(\lambda)+\frac{1}{n+1}\le\alpha_{\text{hall}}$, where $\hat{R}_{\text{hall}}$ is the empirical calibration loss over $n$ documents (Appendix~\ref{app:guarantee}).

\paragraph{Omission controller.}
Source sentence $u_j$ is surfaced when $\hat{p}_{\text{imp}}(u_j)\ge\tau$ and $\hat{p}_{\text{ncov}}(u_j)\ge\gamma$; let $O_{\tau,\gamma}(X)$ be the surfaced set. The fractional loss is
\[
L_{\text{omit}}(\tau,\gamma,X)=\frac{\lvert O_{\text{true}}(X)\setminus O_{\tau,\gamma}(X)\rvert}{\lvert O_{\text{true}}(X)\rvert},
\]
set to 0 when $O_{\text{true}}(X)=\emptyset$. A fractional loss accounts for wide variation in document length and compression: missing one sentence in a long note should not count as a full document failure (Appendix~\ref{app:binary-vs-fractional}).

\paragraph{Two-dimensional calibration with LTT-FST.}
Because a sentence must pass both gates, the policy is the pair $(\tau,\gamma)$. Calibrating one threshold and then imposing the other is not generally valid: the added gate can remove surfaced true omissions and raise the loss. We therefore calibrate both jointly over a pre-specified grid $\mathcal G\subset[0,1]^2$. For each cell $g=(\tau,\gamma)$ with population risk $R_{\mathrm{omit}}(g)=\mathbb E[L_{\mathrm{omit}}(\tau,\gamma,X)]$, we test $H_g: R_{\mathrm{omit}}(g)>\alpha_{\mathrm{omit}}$ with a finite-sample valid p-value $p_g$ from a bounded-loss (Hoeffding--Bentkus) test (Appendix~\ref{app:omission-validity}). Before observing calibration outcomes, we order cells by $\tau+\gamma$ ascending (ties by $\tau$, then $\gamma$), moving from permissive, high-workload policies toward restrictive ones. FST rejects while $p_{g_j}\le\delta_{\mathrm{omit}}$ and stops at the first non-rejection, giving a certified prefix $\widehat{\mathcal G}=\{g_1,\ldots,g_J\}$. By the LTT fixed-sequence guarantee \citep{angelopoulos-ltt-2025},
\[
\Pr\Big(\sup_{g\in\widehat{\mathcal G}} R_{\mathrm{omit}}(g)\le\alpha_{\mathrm{omit}}\Big)\ge 1-\delta_{\mathrm{omit}},
\]
so any selection within $\widehat{\mathcal G}$ inherits it. \textsc{CARE} deploys $g^*=\argmin_{g\in\widehat{\mathcal G}}\widehat W(g)$, the certified cell with the smallest mean calibration workload (surfaced sentences per document), breaking ties by sequence position; Appendix~\ref{app:ordering-ablation} compares deploying $g_J$ and an alternative ordering. Validity does not require monotone risk along the sequence, although the order affects power; we do not tune it to calibration outcomes. If no cell is certified, \textsc{CARE} falls back to $(0,0)$ and surfaces the full source. Since the smallest attainable p-value is $(1-\alpha_{\mathrm{omit}})^n$, at least 19 calibration documents are needed at $\alpha_{\mathrm{omit}}=0.15$, $\delta_{\mathrm{omit}}=0.05$; all primary experiments ($n\ge86$) certified a nonempty set.

\paragraph{Deployment.}
\textsc{CARE} applies $(\lambda^*,\tau^*,\gamma^*)$ at test time to produce a \emph{risk-annotated summary}: the original output with red flags on likely unsupported summary sentences and blue flags on source sentences judged important and likely omitted.

\section{Experimental Setup}
\label{sec:experiments}

\subsection{Datasets}

We evaluate five summarization settings (Table~\ref{tab:dataset-stats}): ACI-Bench \citep{acibench-2023}, MIMIC-BHC \citep{mimic-iv-2023}, MIMIC-CXR \citep{mimic-cxr-2019}, proprietary discharge summaries (Priv-DS), and SumPubMed \citep{sumpubmed-2021}. They span clinical dialogue, hospital course, radiology, multi-note discharge, and biomedical article summarization. The reference summary is never part of the summarizer or judge input: MIMIC-BHC sources exclude the Brief Hospital Course section, MIMIC-CXR sources contain only the FINDINGS section, ACI-Bench sources are dialogue transcripts, and Priv-DS sources exclude the discharge summary. References are used only offline to construct oracle labels. Calibration/test splits are random at the document level. This makes calibration and test documents exchangeable by construction, but it does not guarantee patient independence. A content-based search found few likely repeat patients ($\le$1.2\% of documents), and removing them changed test risk by at most 0.1 percentage points. Appendix~\ref{app:dataset-description} provides details.

\begin{table}[!htbp]
\centering
\scriptsize
\setlength{\tabcolsep}{3pt}
\renewcommand{\arraystretch}{0.85}
\caption{Dataset statistics. $N$: total documents. \emph{Cal/Test}: calibration/test documents in each 70/30 resplit. \emph{Src/Sum Sent}: mean source/summary sentences per document. \emph{Comp.}: compression ratio. \emph{Halluc.\%} and \emph{Omit\%}: oracle-labeled error rates per sentence.}
\label{tab:dataset-stats}
\resizebox{0.6\textwidth}{!}{
\begin{tabular}{lrrrrrrr}
\toprule
Dataset & $N$ & Cal/Test & Src Sent & Sum Sent & Comp. & Halluc.\% & Omit\% \\
\midrule
ACI-Bench     & 123 & 86/37  & 52.0  & 28.4 & 1.8$\times$  & 4.4 & 11.4 \\
MIMIC-BHC     & 500 & 350/150& 115.3 & 17.1 & 6.7$\times$  & 6.1 & 48.6 \\
MIMIC-CXR     & 500 & 350/150& 7.9   & 5.1  & 1.6$\times$  & 6.6 & 9.3 \\
Priv-DS       & 500 & 350/150& 358.5 & 35.8 & 10.0$\times$ & 8.5 & 61.0 \\
SumPubMed     & 500 & 350/150& 169.3 & 13.1 & 12.9$\times$ & 4.4 & 43.5 \\
\bottomrule
\end{tabular}}
\renewcommand{\arraystretch}{1}
\end{table}

\subsection{Models, Calibration, and Metrics}

\textbf{Models.} The base summarizer is Llama-3.3-70B-Instruct (temperature 0), a strong, cost-efficient open-weight model; alternative summarizers are evaluated in Appendix~\ref{app:summarizer-transfer}. GPT-5 (\texttt{gpt-5-2025-08-07}) provides offline oracle labels with a two-pass propose-and-verify procedure. GPT-5-mini (\texttt{gpt-5-mini-2025-08-07}) is the deployment-time judge; across its five replicates per sentence, Fleiss' $\kappa\ge0.71$ and unanimity is 63--90\% (Appendix~\ref{app:judge-variance}), which measures self-consistency rather than clinical validity. Oracle and judge use provider-default stochastic sampling; all calls use \texttt{max\_tokens}=8000. Appendix~\ref{app:repro} lists snapshots, settings, grids, and seeds.

\textbf{Calibration.} Unless stated otherwise, $\alpha_{\mathrm{hall}}=\alpha_{\mathrm{omit}}=0.15$ and $\delta_{\mathrm{omit}}=0.05$ (95\% confidence in the LTT-FST certification). $\alpha=0.15$ is a representative moderate target, not a clinically prescriptive one; we also sweep $\alpha\in\{0.05,0.10,0.15,0.20,0.25,0.30,0.40,0.50\}$ (Figure~\ref{fig:alpha-sweep}; per-dataset results in Figure~\ref{fig:alpha-sweep-per-dataset}). The hallucination controller searches $\lambda\in\{0,0.01,\dots,1\}$, and the omission controller uses $\mathcal G=\{0,0.05,\dots,1\}^2$ (441 cells). Models, rubrics, grids, the LTT ordering, risk levels, and replication settings are fixed before calibration. We report results over 100 random 70/30 calibration/test resplits (seed \texttt{42}), each recalibrating all thresholds; the thresholds in Appendix~\ref{app:detailed} come from a single canonical split with the same seed. In a new setting, \textsc{CARE} is recalibrated on representative local examples, so thresholds are domain-specific; LTT-FST needs at least 19 calibration documents, and workload decreases as the calibration set grows (Appendix~\ref{app:sample-size}). Code, prompts, and judge scores are available at \url{https://github.com/som-shahlab/CARE}.

\textbf{Choosing $\alpha$ and a scorer.} $\alpha$ encodes clinical risk tolerance and trades it against review workload: lower $\alpha$ suits settings where missed content is costly (e.g., discharge medication plans), higher $\alpha$ settings where a downstream reader still checks the source. The scorer determines the workload needed to reach $\alpha$, not whether risk is controlled. GPT-5-mini reaches the target with the fewest flags; Gemini 2.5 Pro reaches similar recall with more flags; local scorers (Qwen3.5-4B, NLI, \textsc{AlignScore}) keep data on premises at the cost of more flags (Section~\ref{sec:detectors}, Appendices~\ref{app:scorer-agnostic} and~\ref{app:cost}). A practical recipe is to shortlist scorers that meet privacy and cost constraints, calibrate each locally at the chosen $\alpha$, and deploy the one with the lowest calibrated workload.

\textbf{Metrics.} For each controller we report the \textbf{violation estimate} $\widehat V=|\mathcal D_{\mathrm{test}}|^{-1}\sum_{X_i\in\mathcal D_{\mathrm{test}}}L(\theta,X_i)$, i.e., the fraction of test documents with any unflagged hallucination or the mean per-document fraction of true omissions not surfaced; \textbf{flagged sentences}, the mean review workload per document; and \textbf{recall}, pooled sentence-level recall of oracle-labeled errors. Violation is document-level and recall sentence-level, so they are not complements. Bracketed intervals are \emph{descriptive resplit intervals} over the 100 resplits: they measure sensitivity to the split, not population confidence.

\section{Results}
\label{sec:results}

\subsection{Overall Safety and Performance}
\label{sec:main-results}
\vspace{-0.3em}
Table~\ref{tab:main-results} reports results at $\alpha=0.15$. Mean held-out violation is at or below $\alpha$ for both controllers on every dataset; calibrated thresholds are in Appendix~\ref{app:detailed}.

\begin{table*}[h]
\centering
\scriptsize
\setlength{\tabcolsep}{3pt}
\renewcommand{\arraystretch}{0.92}
\caption{
Main results at $\alpha = 0.15$ ($\delta_{\mathrm{omit}}=0.05$), averaged over 100 random 70/30 calibration/test resplits. Both controllers have empirical mean violation at or below the target risk level. Flagged reports the mean number of sentences flagged or surfaced per document, with percentage of total sentences in parentheses. Brackets are descriptive 95\% bootstrap intervals over the 100 resplit-level estimates; they summarize sensitivity to the calibration/test split and are not population confidence intervals.
}
\label{tab:main-results}
\rowcolors{2}{gray!6}{white}
\resizebox{0.92\textwidth}{!}{%
\begin{tabular}{llccc ccc}
\toprule
 & & \multicolumn{3}{c}{\textbf{Hallucination}}
   & \multicolumn{3}{c}{\textbf{Omission}} \\
\cmidrule(lr){3-5} \cmidrule(lr){6-8}
Dataset & $N_{\text{test}}$
& Viol.\% & Flagged & Recall\%
& Viol.\% & Flagged & Recall\% \\
\midrule
ACI-Bench   & 37  & 12.2{\tiny [10.9,13.4]} & 3.7 (13.7\%)  & 87.5
                     & 5.7{\tiny [5.0,6.5]} & 15.2 (29.3\%) & 88.7 \\
MIMIC-BHC   & 150 & 12.5{\tiny [12.1,13.0]} & 2.3 (13.5\%) & 85.7
                     & 10.4{\tiny [10.0,10.7]} & 71.4 (62.0\%) & 89.5 \\
MIMIC-CXR   & 150 & 5.7{\tiny [5.4,6.0]}   & 1.2 (24.4\%) & 82.9
                     & 2.0{\tiny [1.8,2.2]} & 2.1 (26.4\%)  & 90.4 \\
Priv-DS     & 150 & 12.8{\tiny [12.1,13.5]} & 8.1 (22.6\%) & 94.9
                     & 9.1{\tiny [9.0,9.2]} & 235.8 (65.7\%) & 90.9 \\
SumPubMed   & 150 & 6.8{\tiny [6.5,7.1]} & 1.0 (7.7\%)  & 86.2
                     & 7.4{\tiny [7.3,7.6]} & 122.4 (72.2\%) & 92.1 \\
\bottomrule
\end{tabular}}
\rowcolors{2}{white}{white}
\renewcommand{\arraystretch}{1}
\end{table*}

\vspace{-0.3em}
\paragraph{Hallucination control.}
\textsc{CARE} flags 8--24\% of summary sentences while capturing 83--95\% of oracle-labeled hallucinations, with mean violation of 5.7--12.8\%. ACI-Bench varies more across resplits because its test split has only 37 documents.

\vspace{-0.3em}
\paragraph{Omission discovery.}
Mean omission violation is 2.0--10.4\%, and \textsc{CARE} recovers 89--92\% of oracle-labeled omissions. Violation sits well below $\alpha$ because LTT-FST certifies the population risk with probability $1-\delta_{\mathrm{omit}}$ rather than only in expectation. Workload varies by task: \textsc{CARE} surfaces 26.4\% of source sentences on MIMIC-CXR and 29.3\% on ACI-Bench, but 62.0--72.2\% on the longer, more compressed MIMIC-BHC, Priv-DS, and SumPubMed. The stricter binary document-level objective is analyzed in Appendix~\ref{app:binary-vs-fractional}.

\subsection{Comparison Against Calibration Baselines}
\label{sec:baselines}
Table~\ref{tab:baselines-main} compares LTT-FST with reduced-dimensional omission controllers held to the same guarantee, $\Pr(R_{\mathrm{omit}}\le\alpha)\ge1-\delta$, with the same Hoeffding--Bentkus test. \textbf{1D-Imp} calibrates only importance, adapting conformal importance selection \citep{kuwahara2025} to our loss; \textbf{Product} calibrates one threshold on $\hat p_{\mathrm{imp}}\hat p_{\mathrm{ncov}}$; \textbf{Union} calibrates the two gates separately with budgets $(\alpha/2,\delta/2)$. Appendix~\ref{app:baselines} also reports them under scalar CRC (an expected-risk guarantee) and uncalibrated Max-F1 and Fixed-0.5 thresholds.

\begin{table}[!t]
\centering
\scriptsize
\setlength{\tabcolsep}{2.5pt}
\caption{Omission controllers held to the same guarantee ($\alpha=0.15$, $\delta=0.05$, Hoeffding--Bentkus $p$-values), averaged over 100 resplits. Each cell reports mean violation \% / surfaced source sentences (\% of source). \textbf{Bold}: lowest workload. \emph{Partial} calibrates importance alone and then applies an uncalibrated coverage gate; it has no guarantee, and its violation is shown in red. Full results are in Table~\ref{tab:baselines}.}
\label{tab:baselines-main}
\resizebox{0.48\textwidth}{!}{%
\begin{tabular}{l cccc c}
\toprule
Dataset & LTT-FST & 1D-Imp & Product (HB) & Union & Partial \\
\midrule
ACI-Bench & 5.7 / \textbf{29.3} & 6.3 / 57.3 & 6.0 / 30.5 & 0.9 / 64.7 & \textcolor{redflag}{36.5} \\
MIMIC-BHC & 10.4 / \textbf{62.0} & 10.0 / 76.1 & 9.8 / 63.3 & 6.8 / 67.7 & \textcolor{redflag}{49.8} \\
MIMIC-CXR & 2.0 / 26.4 & 7.6 / 61.8 & 5.4 / 19.1 & 5.5 / \textbf{15.4} & \textcolor{redflag}{18.8} \\
Priv-DS & 9.1 / \textbf{65.7} & 8.8 / 72.9 & 7.4 / 69.7 & 6.8 / 71.3 & \textcolor{redflag}{46.0} \\
SumPubMed & 7.4 / 72.2 & 6.9 / 81.3 & 10.1 / \textbf{71.4} & 7.1 / 75.1 & \textcolor{redflag}{46.3} \\
\bottomrule
\end{tabular}}
\end{table}

\textbf{An active second gate must be calibrated jointly.} Calibrating importance alone and then adding an uncalibrated coverage gate (Partial) raises omission violation to 18.8--49.8\%, exceeding $\alpha$ on every dataset. Whenever both gates affect the surfacing decision, their joint effect must enter calibration.

\textbf{Coverage reduces unnecessary review.} All guaranteed methods control risk but differ in workload. Importance-only selection must surface every important sentence, including those already covered, and surfaces $1.1$--$2.3\times$ more source sentences than LTT-FST (57.3\% vs.\ 29.3\% on ACI-Bench). LTT-FST has the lowest workload on ACI-Bench, MIMIC-BHC, and Priv-DS; Union and Product are lower on MIMIC-CXR (15.4\% and 19.1\% vs.\ 26.4\%), and Product is marginally lower on SumPubMed (71.4\% vs.\ 72.2\%). Realized violation differs across methods, so these are not comparisons at matched realized risk. Against an unconstrained post-hoc 2D optimizer with no formal guarantee, LTT-FST needs $1.08$--$2.51\times$ the workload (Table~\ref{tab:fst-vs-2djoint}), the price of certifying through a single fixed sequence, largest on short documents.

\subsection{Comparison with Off-the-Shelf Detectors}
\label{sec:detectors}
Faithfulness detectors are typically deployed at a default or F1-optimal threshold. At the default 0.5, document-level hallucination violation exceeds $\alpha=0.15$ in 5 of 7 detector--dataset settings (15.8--19.6\%; Appendix~\ref{app:scorer-agnostic}), and where it meets the target it can do so by over-flagging (NLI flags 58\% of MIMIC-CXR summary sentences). F1-optimal thresholds exceed the target in 22 of 30 dataset--scorer--controller settings (Table~\ref{tab:external-scorers}). Calibrating the same detectors with \textsc{CARE} keeps hallucination violation near the target (9.7--15.9\%) and omission violation below it (2.2--10.3\%), raising the threshold where the default under-flags and lowering it where it over-flags. Table~\ref{tab:scorers-compact} (full version with recall in Table~\ref{tab:external-scorers-efficiency}) shows that calibration controls risk across scorer families but does not equalize efficiency. The LLM judge reaches the target with the fewest flags at comparable or higher recall.

% BEGIN-COMPACT-SCORERS
\begin{table}[!t]
\centering
\scriptsize
\setlength{\tabcolsep}{2.5pt}
\caption{Review workload by scorer at $\alpha=0.15$ after \textsc{CARE} calibration (100 resplits). Each cell: \% of summary sentences flagged (hallucination) / \% of source sentences surfaced (omission); \textbf{bold}: lowest per dataset and controller. Top: LLM judges; bottom: external scorers, with importance held fixed to the LLM judge. Validity is in Table~\ref{tab:external-scorers}; counts and recall are in Table~\ref{tab:external-scorers-efficiency}.}
\label{tab:scorers-compact}
\resizebox{0.48\textwidth}{!}{%
\begin{tabular}{l ccccc}
\toprule
Scorer & ACI & BHC & CXR & Priv-DS & PubMed \\
\midrule
GPT-5-mini & \textbf{13.7} / \textbf{29.3} & \textbf{13.5} / \textbf{62.0} & \textbf{24.4} / \textbf{26.4} & \textbf{22.6} / \textbf{65.7} & \textbf{7.7} / \textbf{72.2} \\
Gemini 2.5 Pro & 17.4 / 35.4 & 14.5 / 63.8 & -- & -- & -- \\
Qwen3.5-4B & 100 / 83.5 & 100 / 74.0 & -- & -- & -- \\
\midrule
NLI (DeBERTa) & 48.3 / 56.6 & 61.2 / 75.8 & 40.2 / 36.8 & 74.2 / 71.8 & 49.3 / 81.1 \\
Embed (BERT) & 86.0 / 64.8 & 84.8 / 89.4 & 44.1 / 64.0 & 94.9 / 86.3 & 55.1 / 90.7 \\
AlignScore & 66.3 / 56.8 & 58.7 / 72.7 & 34.9 / 31.9 & 76.8 / 71.6 & 20.2 / 79.3 \\
\bottomrule
\end{tabular}}
\end{table}
% END-COMPACT-SCORERS

\subsection{Safety--Efficiency Tradeoffs}
\label{sec:alpha-sweep}
\begin{figure*}[t!]
\centering
\resizebox{0.8\textwidth}{!}{\includegraphics{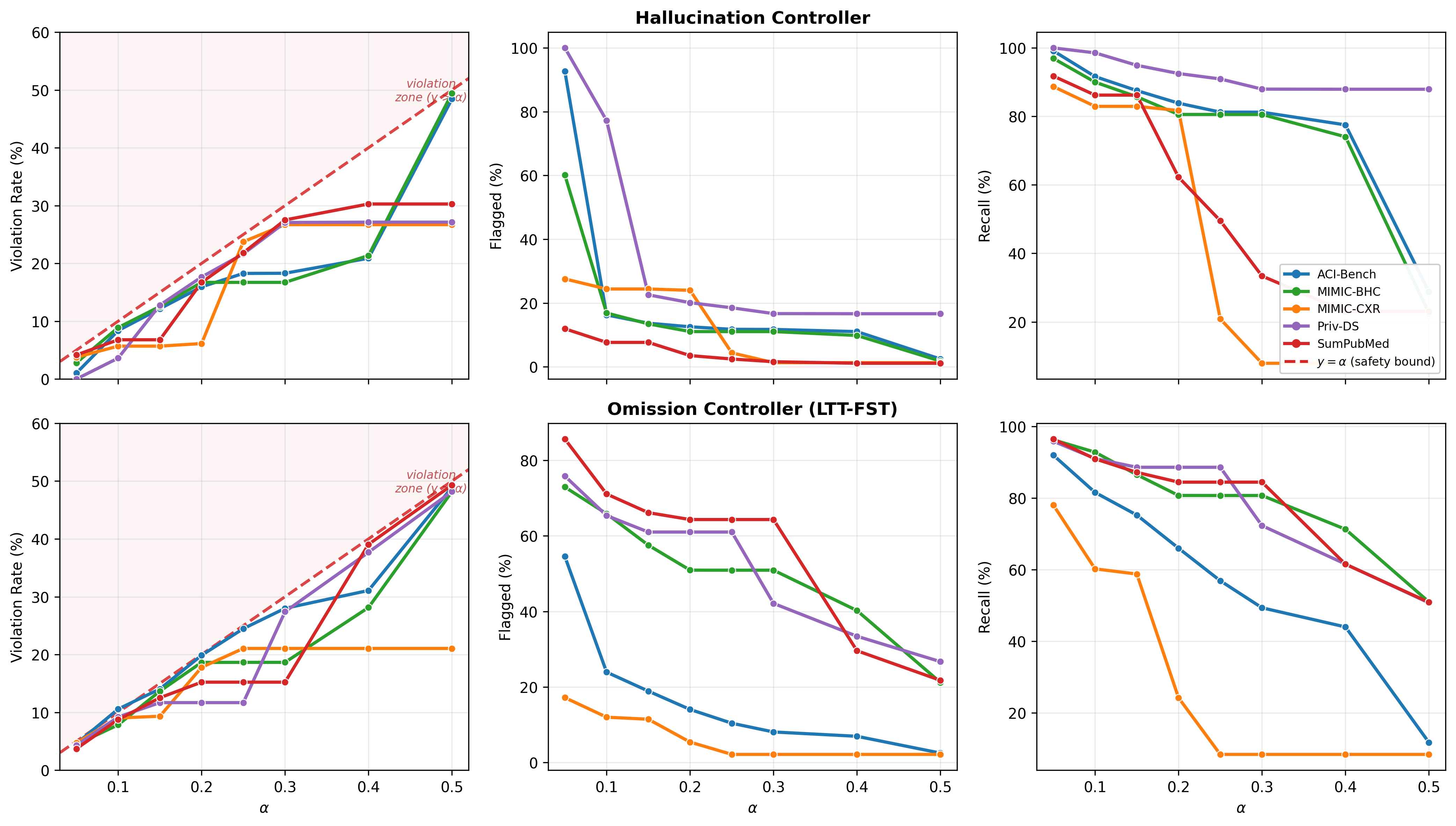}}
\caption{Safety--efficiency tradeoff as $\alpha$ varies. Top: hallucination controller. Bottom: omission controller. Left: violation rate with dashed target line $y=\alpha$. Middle: flagged sentences. Right: recall.}
\label{fig:alpha-sweep}
\end{figure*}

As $\alpha$ increases, \textsc{CARE} flags fewer sentences and allows higher violation (Figure~\ref{fig:alpha-sweep}). Across all datasets and $\alpha \in [0.05,0.50]$, mean violation stays at or below $y=\alpha$ for both controllers.

\subsection{Robustness and Transfer}
\label{sec:robustness}

\textbf{Clinician-defined risk.} We applied the calibrated thresholds unchanged to the 46 clinician-annotated documents and rescored both losses against each clinician's labels. The hallucination controller flags 88\% of clinician-identified hallucinated sentences (73/83; 81--100\% per dataset and clinician). Omission loss under clinician labels stays below $\alpha$ on MIMIC-BHC (9.9\% and 9.7\%, vs.\ 7.6\% under oracle labels); on ACI-Bench it is 6.7\% for one clinician and 21.0\% for the other (3.3\% under oracle labels), mirroring their lower agreement on what counts as an important omission in short encounter notes. With 15--16 documents per dataset these estimates are imprecise (Appendices~\ref{app:error-class} and~\ref{app:clinician-labels}).

\textbf{Oracle-label errors.} When synthetic errors are injected into the hallucination calibration labels, violation rises gradually; on ACI-Bench it stays at or below $\alpha$ with up to 20\% of hallucinations missed. Errors correlated with the judge's own mistakes are hardest, raising violation to 16--17\% on ACI-Bench and MIMIC-BHC and 27\% on Priv-DS (Appendix~\ref{app:oracle-noise}).

\textbf{Prompts, summarizers, and shift.} After recalibration, generic and domain-specific judge prompts give similar risk profiles with dataset-dependent workload (Appendix~\ref{app:prompt-ablation}). Replacing Llama-3.3-70B with Gemini 2.5 Pro, GPT-5, or GPT-5-mini gives hallucination violation of 11.1--14.5\% and omission violation of 4.9--9.7\% (Appendix~\ref{app:summarizer-transfer}). Cross-domain and document-length transfer without recalibration is less reliable (Appendix~\ref{app:distribution-shift}).

\subsection{Sentence-Level Flagging vs.\ Document-Level Abstention}
\label{sec:sent-vs-doc}

Conformal abstention \citep{abbasi-yadkori-abstention-2024,quach-clm-2024} controls the same document-level hallucination risk by rejecting whole summaries. Calibrated with the same CRC procedure, abstention rules that aggregate sentence support by its minimum or mean are both valid, but they reject 28--84\% of summaries, including many with no hallucination. \textsc{CARE} instead flags 8--24\% of summary sentences and keeps the draft (Appendix~\ref{app:sent-vs-doc}): the difference is efficiency and actionability, not validity.

\subsection{Preliminary Assisted-Review Evaluation}
\label{sec:clinician-eval}
In a preliminary randomized within-reader study, three clinicians completed 75 reviews of ACI-Bench and MIMIC-BHC documents (control 38, \textsc{CARE} 37), highlighting important source content missing from the summary and editing it. The primary endpoint, \emph{assisted review-time omission coverage}, is the fraction of oracle-labeled true omissions highlighted by the clinician or surfaced by \textsc{CARE}; it measures what reaches the reviewer's attention, not final summary quality (Table~\ref{tab:clinician-main}).

\begin{table}[!htbp]
\centering
\scriptsize
\setlength{\tabcolsep}{4pt}
\caption{Clinician study decomposition (75 reviews). Percentages are of oracle-labeled true omissions, as means over reviews. $p$-values are from one-sided permutation tests that respect the study design.}
\label{tab:clinician-main}
\begin{tabular}{l rr}
\toprule
Metric & Control & \textsc{CARE} \\
\midrule
Assisted review-time coverage & 50.4\% & \textbf{79.0\%} ($p<0.001$) \\
Clinicians' own highlights & 50.4\% & 60.3\% ($p=0.08$) \\
\textsc{CARE} flags alone & -- & 56.7\% \\
Found by clinicians beyond flags & -- & 22.3\% \\
Minutes per review & 11.4 & 11.3 \\
\bottomrule
\end{tabular}
\end{table}

Assisted coverage rose by 28.6 points with no increase in review time. Clinicians' own detection rose by 9.9 points, with gains for all three clinicians, but not significantly. Clinicians found 22.3\% of true omissions beyond \textsc{CARE}'s flags, so assistance complemented rather than replaced their search, and they independently highlighted 52.2\% of \textsc{CARE}'s omission flags (Appendix~\ref{app:clinician-study}). Deployment cost also varies by task: on Priv-DS, the longest inputs, estimated API cost is \$0.87 per document, mostly importance scoring (about 557K tokens; Appendix~\ref{app:cost}).

\section*{Limitations}
\vspace{-0.5em}

\textsc{CARE}'s guarantees assume calibration data representative of deployment; distribution shift can degrade risk control, as our cross-domain stress tests show, so a new setting requires recalibration \citep{bedi2025fidelity}. The guarantees are marginal, not per-document or subgroup-conditional. They are also defined with respect to the oracle labels. GPT-5 importance labels are anchored to a single reference summary, one notion of what should be retained. Our clinician-label analyses are small (15--16 documents per dataset, mostly from the calibration split) and cover only ACI-Bench, MIMIC-BHC, and MIMIC-CXR; on ACI-Bench, omission risk under clinician labels exceeds the oracle-based estimate. Oracle and judge share a model family, so their errors may be correlated: judge-correlated oracle errors are the case most likely to erode risk control, and the shared family may explain part of the LLM judge's efficiency advantage. Calibrating directly on clinician labels removes this dependence.

Review burden varies by task: on MIMIC-BHC, Priv-DS, and SumPubMed, the omission target requires surfacing 62--72\% of source sentences, so calibration alone cannot ensure a lightweight workflow when the scorer separates omitted from non-omitted content poorly. The fractional guarantee leaves at least one omission unsurfaced in 56--94\% of long documents (Appendix~\ref{app:binary-vs-fractional}), and in copy-forward-heavy Priv-DS records about 19\% of surfaced sentences repeat content already surfaced (Appendix~\ref{app:copy-forward}). LTT-FST cannot certify any cell with fewer than 19 calibration documents, its workload falls as calibration data grow, certifying through one fixed sequence costs up to $2.5\times$ the workload of an unconstrained optimizer on short documents, and the ordering affects power (Appendix~\ref{app:ordering-ablation}). Splits are at the document level and do not guarantee patient independence. The clinician study is small (three readers, 75 reviews); the gain in clinicians' own detection was not significant, final edited summaries were not clinician-scored (automated re-labeling showed similar post-edit omission reductions in both arms), clinicians edited flagged and unflagged hallucinated sentences at similar rates, and flag-level accept/reject actions were not logged.

\paragraph{Future work.} The next step is prospective validation within a health-system summarization workflow, measuring correction quality, false-positive burden, trust, over-reliance, and flag-level feedback, alongside broader clinician validation (including Priv-DS and SumPubMed) and patient-grouped evaluation. \textsc{CARE} could also act as an independent check on agentic or self-verifying summarizers. Methodological extensions include severity-weighted losses, workload-aware fixed-sequence orderings, redundancy-aware displays, and conditional guarantees by document type.

% ============================================================================
\section{Conclusion}
\label{sec:conclusion}

\textsc{CARE} is a post-hoc risk-control layer for LLM-generated medical summaries that combines sentence-level hallucination flagging (scalar CRC) with omission surfacing through a two-dimensional importance--coverage policy calibrated by LTT-FST. Across five tasks, both controllers keep mean held-out violation at or below the target. Calibrating importance and coverage jointly is necessary for validity, coverage reduces review relative to importance-only conformal selection, and sentence-level flagging preserves drafts that document-level abstention would discard. Calibrated thresholds largely carry over to clinician-defined hallucinations, and in a preliminary study \textsc{CARE} increased assisted review-time omission coverage without adding review time. Long, high-compression inputs still require substantial review, which marks where better scorers are most needed.

% ============================================================================
% References
% ============================================================================
\bibliography{references}

% ============================================================================
% Appendix
% ============================================================================
\appendix
\makeatletter
% Appendix \section headings read "Appendix A  Title" (display only).
% \thesection stays "A", so the in-text "Appendix~\ref{...}" cross-references
% remain correct (no doubled "Appendix"), subsections keep "A.1" numbering.
\renewcommand\@seccntformat[1]{%
  \ifnum\pdfstrcmp{#1}{section}=\z@ Appendix~\fi
  \csname the#1\endcsname\quad}
\makeatother

\section{Oracle Label Validation}
\label{app:oracle-validation}

\textsc{CARE} uses LLM-generated oracle labels from Phase~1 as reference labels for risk-controlled calibration. To assess the quality of these labels, two clinicians independently annotated a random sample of documents from ACI-Bench (15 documents, 1{,}291 sentences), MIMIC-BHC (16 documents, 2{,}036 sentences), and MIMIC-CXR (15 documents, 199 sentences). Clinicians annotated all three labeling tasks: factual support, medical importance, and coverage.

We report oracle--human F1 (OH F1), computed as the average of F1(Oracle, $A_1$) and F1(Oracle, $A_2$), following the evaluation paradigm used in SQuAD \citep{rajpurkar-etal-2016-squad}. Human--human F1 (HH F1 = F1($A_1$, $A_2$)) provides an empirical reference point for inter-annotator agreement on each task.

\begin{table}[!htbp]
\centering
\scriptsize
\setlength{\tabcolsep}{3pt}
\renewcommand{\arraystretch}{0.82}
\caption{Oracle label validation against clinician annotations. Two clinicians independently annotated 15--16 randomly sampled documents per dataset. \emph{HH F1}: inter-annotator F1. \emph{OH}: oracle--human F1, precision, and recall, averaged over both annotators. $n$: sentences evaluated.}
\label{tab:oracle-validation}
\resizebox{0.48\textwidth}{!}{
\begin{tabular}{ll r r ccc}
\toprule
 & & & & \multicolumn{3}{c}{\textbf{Oracle--Human}} \\
\cmidrule(lr){5-7}
Dataset & Task & $n$ & HH F1 & F1 & Prec & Rec \\
\midrule
  \multirow{3}{*}{ACI-Bench} & Hallucination & 465 & 0.99 & 0.97 & 0.97 & 0.98 \\
   & Importance & 826 & 0.91 & 0.91 & 0.92 & 0.89 \\
   & Coverage & 826 & 0.91 & 0.91 & 0.90 & 0.93 \\
\midrule
  \multirow{3}{*}{MIMIC-BHC} & Hallucination & 228 & 0.98 & 0.97 & 0.98 & 0.96 \\
   & Importance & 1808 & 0.94 & 0.88 & 0.89 & 0.88 \\
   & Coverage & 1808 & 0.92 & 0.90 & 0.93 & 0.86 \\
\midrule
  \multirow{3}{*}{MIMIC-CXR} & Hallucination & 79 & 0.99 & 0.97 & 0.98 & 0.96 \\
   & Importance & 120 & 0.98 & 0.94 & 0.93 & 0.95 \\
   & Coverage & 120 & 0.97 & 0.95 & 0.95 & 0.95 \\
\bottomrule
\end{tabular}}
\end{table}

Table~\ref{tab:oracle-validation} shows that oracle--human agreement is generally close to human--human agreement across all nine task--dataset combinations. For factual support, oracle--human F1 is 0.97 across all three datasets, close to the human--human F1 of 0.98--0.99. For importance and coverage, oracle--human F1 ranges from 0.88 to 0.95, again close to the corresponding human agreement levels.

Agreement is highest on MIMIC-CXR (OH F1 $\ge$ 0.94), reflecting the structured, formulaic nature of radiology reports. MIMIC-BHC shows the largest gap between oracle--human and human--human agreement for importance and coverage (OH F1 = 0.88--0.90 vs.\ HH F1 = 0.92--0.94), consistent with the difficulty of judging medical relevance in long discharge summaries.

\paragraph{Positive classes.} Table~\ref{tab:oracle-validation} computes F1 with label $=1$ as the positive class for every task: supported, important, and covered. For factuality, about 95\% of annotated summary sentences are supported, so these F1 values mainly reflect agreement on supported content. Error-class agreement is more informative, and we report it next for the exact labels that \textsc{CARE}'s losses use.

\subsection{Agreement on Error Classes and Controlled Labels}
\label{app:error-class}

Table~\ref{tab:error-class} reports sentence-level agreement with each error class as the positive class. It includes the joint omission label, important $\wedge$ not covered, which defines $O_{\mathrm{true}}$. For the joint omission label, oracle--clinician F1 is 0.80 on MIMIC-BHC (clinician--clinician 0.85) and 0.66 on ACI-Bench (0.72). Document-bootstrap 95\% intervals are [0.73, 0.86] vs.\ [0.74, 0.92] on MIMIC-BHC and [0.49, 0.81] vs.\ [0.51, 0.87] on ACI-Bench. MIMIC-CXR has only four omissions per annotator, so its estimates are not informative. Sentence-level hallucinations are rare, with 2--32 positives per annotator and dataset. Agreement on them is moderate for both oracle--clinician pairs (F1 0.31--0.56) and clinician pairs (0.44--0.80). On ACI-Bench, the oracle is more lenient than clinicians (recall 0.48).

\begin{table}[!htbp]
\centering
\scriptsize
\setlength{\tabcolsep}{2.5pt}
\caption{Sentence-level agreement with the error class as positive. \emph{Pos.}: positives labeled by oracle / clinician 1 / clinician 2. \emph{CC}: clinician--clinician F1 (Cohen's $\kappa$). \emph{OC}: oracle--clinician F1, precision, and recall averaged over both clinicians ($\kappa$). \emph{Omission} is the joint label important $\wedge$ not covered.}
\label{tab:error-class}
\resizebox{0.55\textwidth}{!}{%
\begin{tabular}{ll r r c c}
\toprule
Dataset & Positive class & $n$ & Pos. (O/C1/C2) & CC F1 ($\kappa$) & OC F1 / P / R ($\kappa$) \\
\midrule
\multirow{4}{*}{ACI-Bench} & Hallucination & 465 & 22/32/28 & 0.80 (0.79) & 0.56 / 0.66 / 0.48 (0.53) \\
 & Important & 826 & 183/185/196 & 0.91 (0.88) & 0.91 / 0.92 / 0.89 (0.88) \\
 & Not covered & 826 & 452/461/467 & 0.93 (0.85) & 0.93 / 0.94 / 0.92 (0.84) \\
 & \textbf{Omission} & 826 & 27/27/31 & 0.72 (0.71) & 0.66 / 0.69 / 0.64 (0.65) \\
\midrule
\multirow{4}{*}{MIMIC-BHC} & Hallucination & 228 & 14/12/6 & 0.44 (0.42) & 0.48 / 0.39 / 0.67 (0.46) \\
 & Important & 1808 & 574/591/579 & 0.94 (0.90) & 0.88 / 0.89 / 0.88 (0.83) \\
 & Not covered & 1808 & 1344/1300/1312 & 0.97 (0.88) & 0.96 / 0.95 / 0.98 (0.86) \\
 & \textbf{Omission} & 1808 & 279/264/259 & 0.85 (0.82) & 0.80 / 0.78 / 0.83 (0.77) \\
\midrule
\multirow{4}{*}{MIMIC-CXR} & Hallucination & 79 & 4/3/2 & 0.80 (0.79) & 0.31 / 0.25 / 0.42 (0.28) \\
 & Important & 120 & 47/46/46 & 0.98 (0.96) & 0.94 / 0.93 / 0.95 (0.89) \\
 & Not covered & 120 & 21/20/22 & 0.86 (0.83) & 0.79 / 0.79 / 0.79 (0.74) \\
 & \textbf{Omission} & 120 & 4/4/4 & 0.75 (0.74) & 0.50 / 0.50 / 0.50 (0.48) \\
\bottomrule
\end{tabular}}
\end{table}

Table~\ref{tab:confusion} gives the underlying confusion counts, so that error-class precision, recall, and prevalence can be recomputed directly.

\begin{table}[!htbp]
\centering
\scriptsize
\setlength{\tabcolsep}{2.5pt}
\caption{Confusion counts (TP/FP/FN/TN) with the error class as positive. In the oracle columns, the clinician is the reference and the oracle is the prediction. In the C1--C2 column, clinician 1 is the reference.}
\label{tab:confusion}
\resizebox{0.55\textwidth}{!}{%
\begin{tabular}{ll r ccc}
\toprule
Dataset & Positive class & $n$ & Oracle vs.\ C1 & Oracle vs.\ C2 & C2 vs.\ C1 \\
\midrule
\multirow{4}{*}{ACI-Bench} & Hallucination & 465 & 16/6/16/427 & 13/9/15/428 & 24/4/8/429 \\
 & Important & 826 & 170/13/15/628 & 168/15/28/615 & 173/23/12/618 \\
 & Not covered & 826 & 434/18/27/347 & 417/35/50/324 & 433/34/28/331 \\
 & Omission & 826 & 19/8/8/791 & 18/9/13/786 & 21/10/6/789 \\
\midrule
\multirow{4}{*}{MIMIC-BHC} & Hallucination & 228 & 6/8/6/208 & 5/9/1/213 & 4/2/8/214 \\
 & Important & 1808 & 529/45/62/1172 & 495/79/84/1150 & 547/32/44/1185 \\
 & Not covered & 1808 & 1277/67/23/441 & 1274/70/38/426 & 1264/48/36/460 \\
 & Omission & 1808 & 226/53/38/1491 & 207/72/52/1477 & 222/37/42/1507 \\
\midrule
\multirow{4}{*}{MIMIC-CXR} & Hallucination & 79 & 1/3/2/73 & 1/3/1/74 & 2/0/1/76 \\
 & Important & 120 & 44/3/2/71 & 43/4/3/70 & 45/1/1/73 \\
 & Not covered & 120 & 17/4/3/96 & 16/5/6/93 & 18/4/2/96 \\
 & Omission & 120 & 2/2/2/114 & 2/2/2/114 & 3/1/1/115 \\
\bottomrule
\end{tabular}}
\end{table}

\paragraph{Document-level hallucination agreement.} The hallucination loss is defined at the document level, as an indicator that at least one unsupported sentence is left unflagged. We therefore also compare document-level labels, where a document is positive if it contains at least one unsupported sentence (Table~\ref{tab:doc-hall-agreement}). Oracle--clinician agreement is comparable to clinician--clinician agreement on all three datasets.

\begin{table}[!htbp]
\centering
\scriptsize
\caption{Document-level hallucination agreement: documents on which the two labelers agree about whether at least one hallucinated sentence is present.}
\label{tab:doc-hall-agreement}
\begin{tabular}{l c c c}
\toprule
Dataset & Oracle--C1 & Oracle--C2 & C1--C2 \\
\midrule
ACI-Bench (15 docs) & 14/15 & 13/15 & 14/15 \\
MIMIC-BHC (16 docs) & 13/16 & 11/16 & 10/16 \\
MIMIC-CXR (15 docs) & 12/15 & 13/15 & 14/15 \\
\bottomrule
\end{tabular}
\end{table}

\subsection{Calibrated Thresholds Scored Against Clinician Labels}
\label{app:clinician-labels}

To test whether risk calibrated against oracle labels carries over to clinician-defined errors, we applied the canonical-split thresholds (Table~\ref{tab:thresholds}) unchanged to the judge scores of the 46 annotated documents. We then computed both losses three times, using oracle, clinician 1, and clinician 2 labels (Table~\ref{tab:clinician-labels}). Summary sentences were matched to annotation spreadsheets after text normalization. All fuzzy matches were inspected, and only 33 ACI-Bench sentences, 32 of them empty section headers, remained unmatched.

\begin{table}[!htbp]
\centering
\scriptsize
\setlength{\tabcolsep}{3pt}
\caption{\textsc{CARE} thresholds evaluated against oracle / clinician 1 / clinician 2 labels on the annotated documents. \emph{Flagged}: share of labeled hallucinated sentences flagged. \emph{Doc miss}: share of documents with at least one unflagged hallucination. \emph{Omission}: mean fractional omission loss. Most annotated documents fall in the canonical calibration split (11/15, 9/16, 9/15).}
\label{tab:clinician-labels}
\resizebox{0.53\textwidth}{!}{%
\begin{tabular}{l ccc}
\toprule
Dataset & Halluc.\ flagged & Doc miss & Omission \\
\midrule
ACI-Bench & 95\% / 91\% / 81\% & 7 / 20 / 27\% & 3.3 / 6.7 / 21.0\% \\
MIMIC-BHC & 93\% / 83\% / 100\% & 6 / 12 / 0\% & 7.6 / 9.9 / 9.7\% \\
MIMIC-CXR & 75\% / 100\% / 100\% & 7 / 0 / 0\% & 0.0 / 13.3 / 3.3\%$^\ast$ \\
\midrule
Pooled (clinician labels) & 88\% (73/83) & 11\% (C1) / 9\% (C2) & -- \\
\bottomrule
\end{tabular}}
\par\vspace{2pt}\raggedright\footnotesize $^\ast$Only four true omissions per annotator; not interpretable.
\end{table}

The hallucination controller flags 81--100\% of clinician-identified hallucinated sentences, comparable to its recall on oracle-identified ones. Pooled over the 46 documents, the document-level miss rate under clinician labels is 9--11\%. On MIMIC-BHC, omission loss under clinician labels (9.7--9.9\%) is close to oracle-based loss (7.6\%) and below $\alpha$. On ACI-Bench, clinician-labeled omission loss is 6.7\% for clinician 1 but 21.0\% for clinician 2, consistent with the clinicians' lower agreement with each other on the joint omission label. These samples are small, and most documents come from the calibration split. That overlap mainly favors the oracle column; the clinician columns still test whether thresholds transfer to a different labeler.

\subsection{Sensitivity to Oracle-Label Errors}
\label{app:oracle-noise}

We stress-tested the hallucination controller by corrupting only the calibration labels, recalibrating $\lambda$ with the same CRC procedure, and evaluating on the clean (oracle) test labels over the same 100 resplits (Table~\ref{tab:oracle-noise}). Noise models (a)--(e) were fixed before any noisy run. The two added later are labeled post hoc: noise structured after the observed oracle--clinician disagreement, and an $\alpha'$ correction. This analysis measures sensitivity to \emph{additional} errors relative to the oracle, not the oracle's error against clinical truth, which Appendix~\ref{app:clinician-labels} measures directly. It covers the hallucination controller only.

\begin{table}[!htbp]
\centering
\scriptsize
\setlength{\tabcolsep}{3pt}
\caption{Hallucination violation (\%, clean labels) after calibrating on corrupted labels, mean over 100 resplits ($\alpha=0.15$). \emph{Missed $m$}: each hallucination label independently flipped to supported with probability $m$. \emph{Correlated}: misses concentrated on sentences the judge scores as supported (pessimistic). $^\ddagger$Post hoc.}
\label{tab:oracle-noise}
\resizebox{0.6\textwidth}{!}{%
\begin{tabular}{l ccccc}
\toprule
Noise model & ACI & BHC & CXR & Priv-DS & PubMed \\
\midrule
None & 12.2 & 12.5 & 5.7 & 12.8 & 6.8 \\
Missed, $m=0.1$ & 13.0 & 14.0 & 5.7 & 14.2 & 7.8 \\
Missed, $m=0.2$ & 14.7 & 15.8 & 5.9 & 15.9 & 12.4 \\
Missed, $m=0.3$ & 15.9 & 16.7 & 9.0 & 18.0 & 16.9 \\
Missed, $m=0.5$ & 17.9 & 16.7 & 25.8 & 23.9 & 24.1 \\
Correlated worst case & 17.3 & 16.2 & 11.8 & 27.1 & 6.6 \\
Structured after observed disagreement$^\ddagger$ & 11.9--16.7 & 12.5--16.7 & 5.7--17.5 & 20.6--21.8 & 6.8--11.5 \\
$\alpha'=\alpha(1-m)$, missed $m=0.5$$^\ddagger$ & 10.3 & 12.2 & 5.7 & 11.7 & 7.0 \\
\bottomrule
\end{tabular}}
\end{table}

Spurious hallucination labels can only make \textsc{CARE} more conservative. On the annotated documents, oracle false positives fall on borderline sentences that \textsc{CARE} already flags. Missed hallucinations raise violation gradually because the loss is document-level and many hallucinated documents contain more than one hallucination. Misses correlated with the judge are the hardest case, which is plausible because the oracle and judge share a model family. In the observed data, 7 of 19 ACI-Bench oracle misses received a judge score of 1.0. Priv-DS, which has no clinician validation, is most sensitive. The $\alpha'$ correction is valid only under independent misses with a known rate, so we report it for completeness rather than as a recommended fix. These results motivate calibrating directly on clinician-adjudicated labels when they are available.

These results support the use of GPT-5 oracle labels as calibration targets, while recognizing that the resulting guarantees control oracle-defined errors rather than unobserved clinical ground truth.

\subsection{Stratified Resample on Harder Cases}
\label{app:oracle-validation-stratified}

The validation above samples sentences uniformly. To test whether oracle quality is overly optimistic on easier cases, we also evaluate a hard-heavy stratified bootstrap with $N{=}1000$ resamples. In each resample, 50\% of sentences are drawn from cases with non-extreme vote-averaged judge scores, $\hat p\in(0,1)$. These cases include judge disagreement as well as unanimous intermediate judgments and provide a harder-than-random subset for evaluating oracle agreement.

We compare the human--oracle F1 gap under uniform sampling with the gap under the hard-heavy resample (Table~\ref{tab:oracle-validation-stratified}). The change is small across all dataset--task cells ($|\Delta\mathrm{gap}| \le 0.02$), suggesting that oracle agreement does not degrade substantially more than human--human agreement on ambiguous content.

\begin{table}[!htbp]
\centering
\scriptsize
\caption{Stratified oracle validation on harder cases. Each F1 cell reports original / hard-heavy resampled performance. The hard-heavy bootstrap draws 50\% of sentences from cases with non-extreme vote-averaged judge scores ($\hat p\in(0,1)$). \emph{Gap} is Human F1 $-$ Oracle F1, and $\Delta\mathrm{gap}$ is the change from the original sample to the hard-heavy resample.}
\label{tab:oracle-validation-stratified}
\resizebox{0.57\textwidth}{!}{
\begin{tabular}{ll ccc r}
\toprule
Dataset & Task & Oracle F1 & Human F1 & Gap & $\Delta\mathrm{gap}$ \\
 & & \emph{(orig / hard)} & \emph{(orig / hard)} & \emph{(orig / hard)} & \\
\midrule
  ACI-Bench & Factuality & 0.97 / 0.95 & 0.99 / 0.98 & 0.02 / 0.04 & +0.02 \\
   & Importance & 0.91 / 0.89 & 0.91 / 0.89 & 0.00 / 0.01 & +0.01 \\
   & Coverage & 0.91 / 0.90 & 0.91 / 0.91 & 0.00 / 0.01 & +0.01 \\
\midrule
  MIMIC-BHC & Factuality & 0.97 / 0.94 & 0.98 / 0.95 & 0.01 / 0.01 & 0.00 \\
   & Importance & 0.88 / 0.89 & 0.94 / 0.94 & 0.06 / 0.05 & $-0.01$ \\
   & Coverage & 0.90 / 0.87 & 0.91 / 0.90 & 0.01 / 0.03 & +0.02 \\
\midrule
  MIMIC-CXR & Factuality & 0.97 / 0.96 & 0.99 / 0.99 & 0.02 / 0.02 & 0.00 \\
   & Importance & 0.94 / 0.93 & 0.98 / 0.98 & 0.04 / 0.04 & 0.00 \\
   & Coverage & 0.95 / 0.92 & 0.97 / 0.96 & 0.02 / 0.04 & +0.02 \\
\bottomrule
\end{tabular}}
\end{table}

\section{Formal Risk-Control Guarantees}
\label{app:guarantee}

The main text adopts i.i.d. calibration and deployment documents as a common sufficient assumption for both controllers. Scalar CRC itself requires only exchangeability, so we state that weaker condition in the theorem below.

\begin{theorem}[Scalar Conformal Risk Control]
Following \citet{angelopoulos-crc-2024},
Let $\{(X_i,Y_i)\}_{i=1}^{n+1}$ be exchangeable random variables, and let
$L(\lambda,X,Y)\in[0,1]$ be a loss function that is non-increasing in
$\lambda$ for each $(X,Y)$. Define
\begin{align*}
\widehat R^{+}(\lambda)
&=
\frac{\sum_{i=1}^{n} L(\lambda,X_i,Y_i) + 1}{n+1},\\
\lambda^*
&=
\inf\left\{\lambda \in \Lambda :
\widehat R^{+}(\lambda)\le \alpha\right\}.
\end{align*}
Then
\[
\mathbb{E}\left[L(\lambda^*,X_{n+1},Y_{n+1})\right]\le \alpha.
\]
\end{theorem}

The hallucination controller directly instantiates this scalar result with $L=L_{\mathrm{hall}}(\lambda,X)$ and $\lambda\in[0,1]$, searched at resolution 0.01.

The omission controller does not apply the scalar infimum rule. Because its operating point is selected from a two-dimensional parameter grid, we instead use the Learn-Then-Test fixed-sequence procedure described next.

\subsection{LTT-FST Guarantee for 2D Omission Calibration}
\label{app:omission-validity}

For each grid cell $g=(\tau,\gamma)\in\mathcal G$, define
\[
R_g
=
R_{\mathrm{omit}}(\tau,\gamma)
=
\mathbb E[L_{\mathrm{omit}}(\tau,\gamma,X)]
\]
and the null hypothesis
\[
H_g:R_g>\alpha_{\mathrm{omit}}.
\]

Given $n$ calibration documents, let
\[
L_{i,g}
=
L_{\mathrm{omit}}(\tau,\gamma,X_i),
\qquad
\widehat R_g
=
\frac{1}{n}\sum_{i=1}^n L_{i,g}.
\]
% Since $L_{i,g}\in[0,1]$, we use the finite-sample Hoeffding--Bentkus p-value of \citet{angelopoulos-ltt-2025}:
% \[
% p_g^{\mathrm{HB}}
% =
% \min\left\{
% \exp\!\left[
% -n\,h_1(
% \widehat R_g\wedge\alpha_{\mathrm{omit}},
% \alpha_{\mathrm{omit}}
% )
% \right],
% \,
% e\,\Pr\!\left[
% \operatorname{Bin}(n,\alpha_{\mathrm{omit}})
% \le
% \left\lceil n\widehat R_g\right\rceil
% \right]
% \right\},
% \]
% where
% \[
% h_1(a,b)
% =
% a\log\frac{a}{b}
% +
% (1-a)\log\frac{1-a}{1-b}.
% \]
% Under $H_g$, $p_g^{\mathrm{HB}}$ is super-uniform and therefore provides a valid p-value for testing whether the population risk exceeds $\alpha_{\mathrm{omit}}$.

Since $L_{i,g}\in[0,1]$, we use the finite-sample
Hoeffding--Bentkus p-value of \citet{angelopoulos-ltt-2025}:
\[
\begin{aligned}
p_g^{\mathrm{HB}}
=\min\Biggl\{&
\exp\!\left[
-n\,h_1\!\left(
\widehat R_g\wedge\alpha_{\mathrm{omit}},
\alpha_{\mathrm{omit}}
\right)
\right],
\\
&
e\,\Pr\!\left[
\operatorname{Bin}(n,\alpha_{\mathrm{omit}})
\le \left\lceil n\widehat R_g\right\rceil
\right]
\Biggr\}.
\end{aligned}
\]
Here,
\[
h_1(a,b)
=
a\log\frac{a}{b}
+
(1-a)\log\frac{1-a}{1-b}.
\]
Under $H_g$, $p_g^{\mathrm{HB}}$ is super-uniform and therefore
provides a valid p-value for testing whether the population risk
exceeds $\alpha_{\mathrm{omit}}$.

\paragraph{Power at small calibration sizes.}
The finite-sample guarantee can be conservative when calibration data are scarce. The smallest attainable Hoeffding--Bentkus p-value occurs at $\widehat R_g=0$ and equals
\[
p_{\min}=(1-\alpha_{\mathrm{omit}})^n.
\]
At $\alpha_{\mathrm{omit}}=0.15$ and $\delta_{\mathrm{omit}}=0.05$, no candidate can be certified for $n<19$, even with zero empirical calibration loss. Small calibration sets can therefore force the conservative fallback below.

\paragraph{Fixed-sequence testing.}
Before observing calibration outcomes, we sort grid cells by $\tau+\gamma$ in ascending order, with ties broken by $\tau$ ascending and then $\gamma$ ascending. This deterministic ordering depends only on the pre-specified grid coordinates. We apply single-start fixed-sequence testing at significance level $\delta_{\mathrm{omit}}$. Beginning with $g_1$, we reject $H_{g_j}$ and continue while
\[
p_{g_j}^{\mathrm{HB}}\le\delta_{\mathrm{omit}},
\]
stopping at the first non-rejection. If $J$ denotes the final rejected index, the certified set is
\[
\widehat{\mathcal G}=\{g_1,\ldots,g_J\},
\]
and \textsc{CARE} deploys
\[
g^*=\argmin_{g\in\widehat{\mathcal G}}\ \widehat W(g),\qquad \widehat W(g)=\frac1n\sum_{i=1}^n \lvert O_{g}(X_i)\rvert,
\]
the certified cell with the smallest mean calibration workload, with ties broken by sequence position. Because the guarantee below holds simultaneously for all certified cells, this choice cannot invalidate it. Selecting by workload within $\widehat{\mathcal G}$ does not require any additional multiplicity correction. All reported primary experiments yielded a nonempty certified set. If no hypothesis is rejected, \textsc{CARE} falls back to $(\tau,\gamma)=(0,0)$ and surfaces every source sentence. Since this policy has $L_{\mathrm{omit}}(0,0,X)=0$ for every document, it provides a deterministic safe fallback.

Fixed-sequence testing controls the family-wise error rate at level $\delta_{\mathrm{omit}}$. Therefore, by \citet[Theorem~1]{angelopoulos-ltt-2025},
\[
\Pr\left(
\sup_{g\in\widehat{\mathcal G}}
R_{\mathrm{omit}}(g)
\le
\alpha_{\mathrm{omit}}
\right)
\ge
1-\delta_{\mathrm{omit}}.
\]
Consequently,
\[
\Pr\left(
R_{\mathrm{omit}}(g^*)
\le
\alpha_{\mathrm{omit}}
\right)
\ge
1-\delta_{\mathrm{omit}}.
\]

\paragraph{Ordering and monotonicity.}
The fixed sequence is a pre-specified priority ordering rather than a monotone path through the two-dimensional parameter space. Adjacent cells may increase one threshold while decreasing the other, and risk need not be monotone along the sequence. Fixed-sequence validity does not require such monotonicity. The particular ordering can nevertheless affect power because a single non-rejection terminates the sequence. Appendix~\ref{app:ordering-ablation} evaluates the deploy rule and an alternative ordering.

Separately, omission loss is coordinatewise monotone:
\[
\begin{aligned}
\tau' \le \tau,\quad \gamma' \le \gamma
\quad &\Longrightarrow \\
L_{\mathrm{omit}}(\tau',\gamma',X)
&\le L_{\mathrm{omit}}(\tau,\gamma,X).
\end{aligned}
\]

Thus, relaxing either gate while holding the other fixed can only surface additional sentences and cannot increase missed-omission loss. The fixed-sequence guarantee does not rely on this property; it explains why the most permissive cells at the start of the sequence have low risk and are certified first.

\paragraph{Unconstrained empirical comparator.}
For comparison only, define
\[
\widehat R_{\mathrm{emp}}^{+}(g)
=
\frac{n}{n+1}\widehat R_g+\frac{1}{n+1}
\]
and
\[
\widehat{\mathcal F}_{\mathrm{emp}}
=
\left\{
g\in\mathcal G:
\widehat R_{\mathrm{emp}}^{+}(g)
\le
\alpha_{\mathrm{omit}}
\right\}.
\]
The 2D-Joint comparator selects the minimum-workload cell from $\widehat{\mathcal F}_{\mathrm{emp}}$. Because this choice optimizes post hoc over the full grid, we use it only as an empirical comparator and do not assign it the LTT-FST guarantee.

Table~\ref{tab:fst-vs-2djoint} compares LTT-FST with 2D Joint on all five datasets over 100 random 70/30 calibration/test resplits. 2D Joint operates close to $\alpha$, reaching 15.0\% on ACI-Bench, whereas LTT-FST operates at 2.0--10.4\%. LTT-FST requires $1.08$--$1.09\times$ the comparator's workload on the three long-document datasets, but $1.58\times$ on ACI-Bench and $2.51\times$ on MIMIC-CXR. On short documents, the high-probability guarantee and the fixed sequence's early stopping leave a larger margin below $\alpha$. This margin is the cost of a formal two-dimensional selection guarantee.

\begin{table}[!htbp]
\centering
\scriptsize
\setlength{\tabcolsep}{3pt}
\renewcommand{\arraystretch}{0.82}
\caption{LTT-FST versus the unconstrained 2D-Joint comparator at $\alpha=0.15$, averaged over 100 random 70/30 calibration/test resplits. LTT-FST uses FWER-controlled fixed-sequence testing, whereas 2D Joint minimizes workload over the full empirical candidate set and has no formal two-dimensional selection guarantee. Ratio denotes LTT-FST workload divided by comparator workload. Brackets are descriptive bootstrap intervals over resplit-level estimates.}
\label{tab:fst-vs-2djoint}
\resizebox{0.48\textwidth}{!}{%
\begin{tabular}{l cc cc r}
\toprule
 & \multicolumn{2}{c}{\textbf{LTT-FST}} & \multicolumn{2}{c}{\textbf{2D Joint (comparator)}} & \\
\cmidrule(lr){2-3} \cmidrule(lr){4-5}
Dataset & V\% [95\% int.] & WL & V\% [95\% int.] & WL & Ratio \\
\midrule
ACI-Bench  & 5.7 [5.0, 6.5]     &  15.2 & 15.0 [13.8, 16.3]  &   9.6 & 1.58$\times$ \\
MIMIC-BHC  & 10.4 [10.0, 10.7]  &  71.4 & 13.9 [13.7, 14.2]  &  65.9 & 1.08$\times$ \\
MIMIC-CXR  & 2.0 [1.8, 2.2]     &   2.1 & 11.6 [10.7, 12.6]  &   0.8 & 2.51$\times$ \\
Priv-DS    & 9.1 [9.0, 9.2]     & 235.8 & 11.7 [11.6, 11.8]  & 219.0 & 1.08$\times$ \\
SumPubMed  & 7.4 [7.3, 7.6]     & 122.4 & 12.5 [12.2, 12.9]  & 112.1 & 1.09$\times$ \\
\bottomrule
\end{tabular}}
\renewcommand{\arraystretch}{1}
\end{table}

\subsection{Deploy Rule and Ordering Ablation}
\label{app:ordering-ablation}

Table~\ref{tab:ordering-ablation} compares three LTT procedures that all carry the same $(\alpha,\delta)$ guarantee. The first deploys the final rejected cell $g_J$ of the $\tau+\gamma$ sequence. The second is \textsc{CARE}'s rule, which deploys the minimum-workload certified cell of the same sequence. The third learns the ordering on a held-out third of the calibration set: candidate cells with empirical risk at most $\alpha$ on that part are ordered by decreasing workload, and fixed-sequence testing runs on the remaining two-thirds. Because the ordering depends only on data disjoint from the certification data, the third procedure is also valid. Its rule was fixed before running.

\begin{table}[!htbp]
\centering
\scriptsize
\setlength{\tabcolsep}{3pt}
\caption{Deploy-rule and ordering ablation at $\alpha=0.15$, $\delta=0.05$, 100 resplits. Each cell reports mean violation \% / mean surfaced source sentences per document / \% of resplits whose test violation exceeds $\alpha$.}
\label{tab:ordering-ablation}
\resizebox{0.53\textwidth}{!}{%
\begin{tabular}{l ccc}
\toprule
Dataset & $\tau+\gamma$, last ($g_J$) & $\tau+\gamma$, min-WL (\textsc{CARE}) & Split-learned order \\
\midrule
ACI-Bench & 4.6 / 23.0 / 3 & 5.7 / 15.2 / 4 & 4.1 / 31.0 / 2 \\
MIMIC-BHC & 8.5 / 74.5 / 0 & 10.4 / 71.4 / 0 & 6.0 / 83.1 / 0 \\
MIMIC-CXR & 0.2 / 7.5 / 0 & 2.0 / 2.1 / 0 & 7.1 / 1.5 / 0 \\
Priv-DS & 8.8 / 261.6 / 0 & 9.1 / 235.8 / 0 & 9.7 / 230.0 / 0 \\
SumPubMed & 8.3 / 138.5 / 0 & 7.4 / 122.4 / 0 & 7.8 / 124.5 / 0 \\
\bottomrule
\end{tabular}}
\end{table}

Deploying the minimum-workload certified cell reduces workload relative to $g_J$ on every dataset, by a factor of $1.04$--$3.6$, at no cost to validity. Learning the ordering on held-out calibration data helps on MIMIC-CXR and Priv-DS but hurts on ACI-Bench and MIMIC-BHC, where setting aside a third of the calibration set for ordering reduces the power of the certification step. No single data-independent ordering dominates. Workload-aware orderings are a promising direction when calibration data are plentiful.

\subsection{Calibration-Set Size}
\label{app:sample-size}

To assess how much labeled data a new deployment needs, we subsample $n\in\{15,25,50,75,100\}$ calibration documents (20 repeats each) and evaluate on a fixed held-out test set (Table~\ref{tab:sample-size}). With $n=15$, no cell can be certified, because $(1-0.15)^{15}>0.05$, so LTT-FST falls back to surfacing the full source. From $n=25$, every repeat certified a policy. Workload then falls steadily as $n$ grows, while violation stays well below $\alpha$. Calibration-set size therefore mainly affects efficiency rather than validity. With roughly 50--100 documents, workload approaches the full-data operating point.

\begin{table}[!htbp]
\centering
\scriptsize
\setlength{\tabcolsep}{3pt}
\caption{LTT-FST omission control vs.\ calibration-set size ($\alpha=0.15$, $\delta=0.05$; 20 subsamples per size). Each cell reports mean violation \% / mean surfaced source sentences per document. At $n=15$ no cell can be certified, and the fallback surfaces every source sentence. ACI-Bench has 86 calibration documents, so $n=100$ is not available.}
\label{tab:sample-size}
\resizebox{0.53\textwidth}{!}{%
\begin{tabular}{l ccccc}
\toprule
Dataset & $n=15$ & $n=25$ & $n=50$ & $n=75$ & $n=100$ \\
\midrule
ACI-Bench & 0.0 / 51.0 & 1.4 / 34.2 & 4.2 / 22.2 & 5.3 / 16.5 & -- \\
MIMIC-BHC & 0.0 / 115.7 & 0.7 / 99.3 & 3.8 / 83.5 & 5.4 / 79.8 & 6.2 / 78.8 \\
MIMIC-CXR & 0.0 / 7.9 & 1.6 / 3.0 & 1.3 / 2.1 & 1.1 / 2.1 & 1.5 / 2.1 \\
Priv-DS & 0.0 / 358.4 & 0.3 / 329.7 & 4.2 / 276.0 & 5.4 / 262.3 & 6.6 / 249.4 \\
SumPubMed & 0.0 / 168.3 & 0.9 / 155.5 & 3.8 / 136.8 & 5.1 / 130.6 & 6.0 / 128.6 \\
\bottomrule
\end{tabular}}
\end{table}

\paragraph{Fractional vs.\ binary omission loss}
Binary loss treats any missed omission as a full document failure. Fractional loss gives partial credit: missing 1 of 20 true omissions incurs loss 0.05 rather than 1.0. Both are bounded losses that can be used with the corresponding risk-control calibration procedure. We use fractional loss because it better reflects documents with many omissions, improves comparability across datasets with different omission counts, and yields more informative calibration in high-compression settings.

\paragraph{Severity-weighted losses}
The framework can also accommodate pre-specified bounded harm- or severity-weighted losses. Let $w(X)\ge0$ be a bounded severity weight and define $L_w(\theta,X)=w(X)L(\theta,X)$, rescaled if necessary so that $L_w\in[0,1]$. Such a loss can be incorporated provided the assumptions of the corresponding risk-control procedure remain satisfied. We use unweighted fractional loss because clinically validated severity weights are unavailable in our datasets.

\begin{figure*}[t!]
    \centering
    \resizebox{0.92\textwidth}{!}{\includegraphics{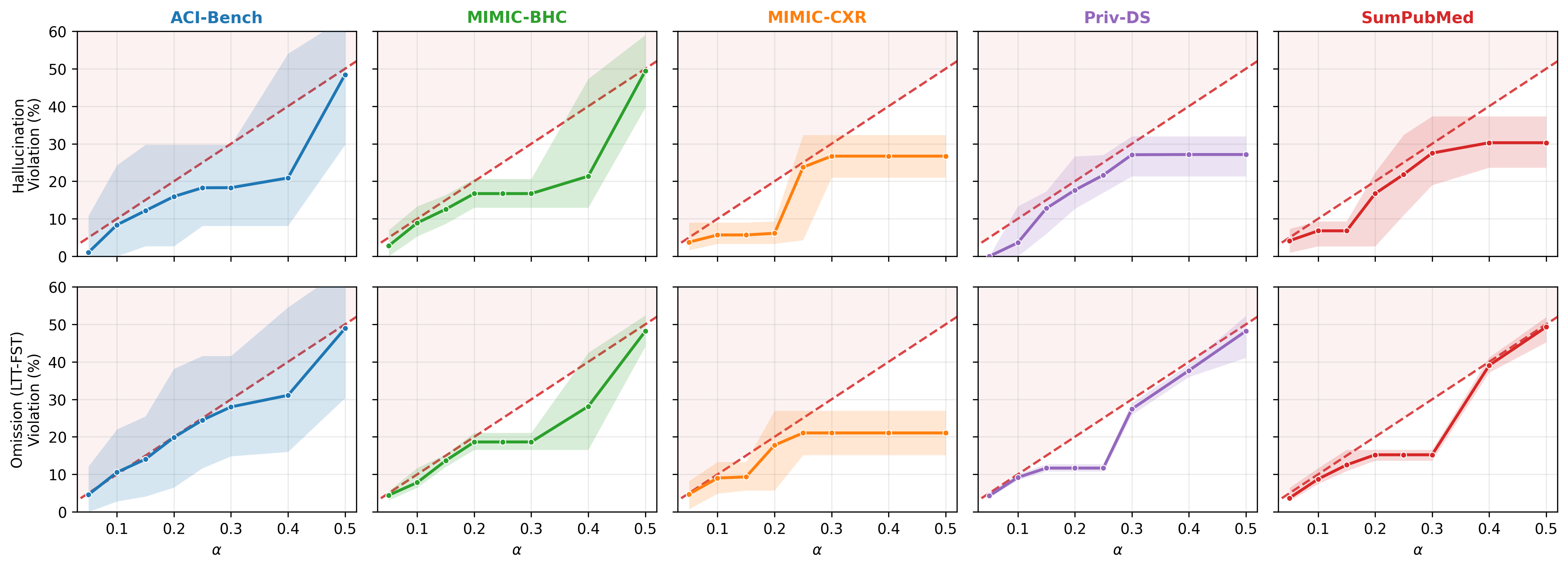}}
    \caption{Per-dataset $\alpha$ sweep over 100 random 70/30 calibration/test resplits. Rows show hallucination control and omission control, columns show datasets. Solid curves report mean violation rate, shaded bands show 2.5--97.5 percentile intervals across resplits, and dashed lines mark $y=\alpha$.}
    \label{fig:alpha-sweep-per-dataset}
\end{figure*}

\paragraph{Empirical validity across $\alpha$}
Figure~\ref{fig:alpha-sweep-per-dataset} disaggregates the main-text $\alpha$ sweep by dataset. Each panel shows the mean violation rate against the target line $y=\alpha$, with 95\% percentile intervals across 100 random 70/30 calibration/test resplits. Across all datasets and $\alpha\in[0.05,0.50]$, the resplit-averaged violation rate remains at or below $y=\alpha$ for both controllers. Realized held-out estimates can exceed $\alpha$ even under a valid calibration procedure. For hallucination control, CRC controls expected loss over calibration and future-sample randomness. For omission control, LTT-FST guarantees with probability at least $1-\delta_{\mathrm{omit}}$ over calibration that the population omission risk is at most $\alpha_{\mathrm{omit}}$. Finite test-set sampling can nevertheless produce an empirical violation estimate above $\alpha_{\mathrm{omit}}$.

\section{Binary vs.\ Fractional Omission Loss}
\label{app:binary-vs-fractional}

Our omission controller uses fractional loss: it bounds the expected fraction of true omissions missed per document, rather than the probability that any omission is missed. Table~\ref{tab:binary-vs-fractional} reports both fractional and binary omission loss for each method.

The distinction matters most in long, high-compression datasets. On ACI-Bench and MIMIC-CXR, documents contain relatively few true omissions, so fractional and binary loss differ by at most 9 percentage points (1--6 points for LTT-FST). In contrast, on MIMIC-BHC, Priv-DS, and SumPubMed, documents often contain many true omissions. In these settings, a method can catch most omissions in each document and still miss at least one omission in most documents. For example, LTT-FST keeps fractional loss at or below $\alpha=0.15$ on all datasets, but binary loss reaches 56--94\% on high-compression datasets.

This does not indicate a failure of the fractional guarantee. It reflects a different deployment objective. Fractional loss is appropriate when the goal is to control overall omission burden while preserving review efficiency. Binary loss is appropriate when any missed omission is unacceptable, but achieving such a guarantee on long documents would require surfacing substantially more source content.

\begin{table}[!htbp]
\centering
\small
\caption{Fractional vs.\ binary omission loss at $\alpha=0.15$, averaged over 100 random calibration/test resplits. Fractional loss measures the mean fraction of true omissions missed per document, binary loss measures the fraction of documents with any missed omission. Bold identifies \textsc{CARE}'s primary omission controller.}
\label{tab:binary-vs-fractional}
\resizebox{0.4\textwidth}{!}{%
\begin{tabular}{ll cc}
\toprule
Dataset & Method & Frac.\% & Binary\% \\
\midrule
  ACI-Bench & \textbf{LTT-FST} & \textbf{5.7} & 11.9 \\
   & 1D-Imp (CRC) & 11.9 & 20.6$^\dagger$ \\
   & Product (CRC) & 14.1 & 22.2$^\dagger$ \\
   & Union (CRC) & 11.3 & 19.8$^\dagger$ \\
\cdashline{2-4}
   & \textcolor{gray}{Partial\textsuperscript{*}} & \textcolor{redflag}{36.5} & \textcolor{gray}{45.5} \\
\midrule
  MIMIC-BHC & \textbf{LTT-FST} & \textbf{10.4} & 63.0$^\dagger$ \\
   & 1D-Imp (CRC) & 11.9 & 65.7$^\dagger$ \\
   & Product (CRC) & 10.1 & 65.7$^\dagger$ \\
   & Union (CRC) & 10.1 & 66.5$^\dagger$ \\
\cdashline{2-4}
   & \textcolor{gray}{Partial\textsuperscript{*}} & \textcolor{redflag}{49.8} & \textcolor{gray}{98.6} \\
\midrule
  MIMIC-CXR & \textbf{LTT-FST} & \textbf{2.0} & 2.6 \\
   & 1D-Imp (CRC) & 7.6 & 8.5 \\
   & Product (CRC) & 5.6 & 6.5 \\
   & Union (CRC) & 8.1 & 9.4 \\
\cdashline{2-4}
   & \textcolor{gray}{Partial\textsuperscript{*}} & \textcolor{redflag}{18.8} & \textcolor{gray}{19.5} \\
\midrule
  Priv-DS & \textbf{LTT-FST} & \textbf{9.1} & 93.6$^\dagger$ \\
   & 1D-Imp (CRC) & 8.8 & 92.7$^\dagger$ \\
   & Product (CRC) & 13.0 & 96.3$^\dagger$ \\
   & Union (CRC) & 9.2 & 95.3$^\dagger$ \\
\cdashline{2-4}
   & \textcolor{gray}{Partial\textsuperscript{*}} & \textcolor{redflag}{46.0} & \textcolor{gray}{99.3} \\
\midrule
  SumPubMed & \textbf{LTT-FST} & \textbf{7.4} & 55.9$^\dagger$ \\
   & 1D-Imp (CRC) & 6.9 & 51.6$^\dagger$ \\
   & Product (CRC) & 10.1 & 69.1$^\dagger$ \\
   & Union (CRC) & 11.9 & 75.1$^\dagger$ \\
\cdashline{2-4}
   & \textcolor{gray}{Partial\textsuperscript{*}} & \textcolor{redflag}{46.3} & \textcolor{gray}{98.9} \\
\bottomrule
\end{tabular}}
\vspace{2pt}
\par\small $^\dagger$Binary loss exceeds $\alpha=0.15$. Baselines are the scalar-CRC variants (Table~\ref{tab:baselines}). $^*$Partial applies an uncalibrated second gate and therefore has no formal risk-control guarantee.
\end{table}

\subsection{Marginal Guarantees and Per-Document Tail Risk}
\label{app:tail-risk}

\textsc{CARE}'s statistical guarantees are marginal and do not imply that every individual document has loss below $\alpha$. Table~\ref{tab:tail-risk} reports empirical percentiles of the per-document fractional omission loss, pooled across all 100 calibration/test resplits for each dataset.

The mean per-document loss is at or below $\alpha=0.15$ on every dataset, consistent with the target risk bound. Tail behavior depends strongly on document structure. Long, high-omission-count datasets such as MIMIC-BHC, Priv-DS, and SumPubMed have smoother tails: their $P95$ losses range from 0.21 to 0.33 because each document contains many true omissions and fractional loss varies more continuously. Short-document datasets such as ACI-Bench and MIMIC-CXR have more discrete tails: most documents have zero loss, but $P99=1.0$. When a document has only one true omission, missing that omission yields fractional loss $L=1$.

These results illustrate the gap between marginal risk control and per-document guarantees. They motivate future work on conditional conformal methods that stratify risk by document type, clinical domain, or document complexity \citep{gibbs2025conditional-conformal,campos-cp-nlp-survey-2024}.

\begin{table}[!htbp]
\centering
\scriptsize
\caption{Per-document tail-risk diagnostics for the omission controller at $\alpha=0.15$. Percentiles are computed from pooled per-document fractional omission losses across 100 calibration/test resplits.}
\label{tab:tail-risk}
\resizebox{0.48\textwidth}{!}{%
\begin{tabular}{l r r r r r r r}
\toprule
Dataset & $n$ & Mean & $P50$ & $P75$ & $P90$ & $P95$ & $P99$ \\
\midrule
  ACI-Bench & 3,700 & 0.057 & 0.000 & 0.000 & 0.167 & 0.500 & 1.000 \\
  MIMIC-BHC & 15,000 & 0.104 & 0.067 & 0.158 & 0.263 & 0.333 & 0.667 \\
  MIMIC-CXR & 15,000 & 0.020 & 0.000 & 0.000 & 0.000 & 0.000 & 1.000 \\
  Priv-DS & 15,000 & 0.091 & 0.078 & 0.121 & 0.172 & 0.213 & 0.303 \\
  SumPubMed & 15,000 & 0.074 & 0.050 & 0.122 & 0.200 & 0.250 & 0.417 \\
\bottomrule
\end{tabular}}
\end{table}

\section{Dataset Description}
\label{app:dataset-description}
Table~\ref{tab:datasets} summarizes the five datasets used in our evaluation. The datasets span dialogue summarization, hospital-course synthesis, radiology report summarization, multi-note discharge summarization, and biomedical article summarization, covering a wide range of document lengths and compression ratios.
\begin{table}[!htbp]
\centering
\scriptsize
\renewcommand{\arraystretch}{0.82}
\caption{Evaluation datasets. The tasks vary in source length, compression ratio, and clinical or biomedical domain.}

\label{tab:datasets}
\setlength{\tabcolsep}{4pt}
\begin{tabular}{>{\raggedright\arraybackslash}p{2.3cm} p{4.9cm}}
\toprule
\textbf{Dataset} & \textbf{Description} \\
\midrule

\textbf{ACI-Bench} \citep{acibench-2023} &
Doctor--patient dialogue summarized into structured visit notes. This dataset tests \textsc{CARE} on conversational clinical documentation, where summaries must preserve clinically relevant details from dialogue. \\
\midrule

\textbf{MIMIC-BHC} \citep{mimic-iv-2023} &
Clinical notes summarized into a Brief Hospital Course. This high-compression task ($6.7\times$) tests whether \textsc{CARE} can surface important omitted content from longitudinal hospitalizations. \\
\midrule

\textbf{MIMIC-CXR} \citep{mimic-cxr-2019} &
Radiology findings summarized into impressions. Reports are short and formulaic, providing a setting with dense technical language and limited score separability. \\
\midrule

\textbf{Priv-DS} &
Proprietary multi-note discharge summarization dataset. Sources include all encounter notes for an admission, with an average of 24.5 notes and 358.5 source sentences per admission. This dataset tests scalability to long, real-world clinical encounters under high compression ($10.0\times$). \\
\midrule

\textbf{SumPubMed} \citep{sumpubmed-2021} &
Biomedical research articles summarized into structured abstracts. This dataset tests \textsc{CARE} outside clinical notes in an extreme compression setting ($12.9\times$). \\

\bottomrule
\end{tabular}
\renewcommand{\arraystretch}{1}
\end{table}

\subsection{Data Use and Privacy Safeguards}
\label{app:data-governance}
MIMIC-BHC and MIMIC-CXR were accessed by credentialed researchers under the PhysioNet Credentialed Health Data Use Agreement. All LLM calls on these data, for every model, were made through a PHI-compliant Microsoft Azure environment that does not retain or train on submitted data, consistent with PhysioNet guidance on responsible LLM use. Priv-DS consists of de-identified encounter notes accessed under an institutional data-use agreement and was processed only within the same PHI-compliant Microsoft Azure environment. Priv-DS cannot be redistributed. ACI-Bench and SumPubMed are public. Local models (Qwen3.5-4B and the external scorers) ran on institutional hardware.

\subsection{Reference Exclusion and Patient Overlap}
\label{app:leakage}
The reference summary is never included in the source passed to the summarizer or judge (Table~\ref{tab:leakage}). For Priv-DS, 22\% of reference sentences also appear near-verbatim in the encounter notes. This reflects routine copy-forward from history-and-physical, imaging, and progress notes, not leakage of the reference: none of the five discharge-titled source notes shares a sentence with the reference.

\begin{table}[!htbp]
\centering
\scriptsize
\setlength{\tabcolsep}{3pt}
\caption{Reference-exclusion checks. \emph{Ref.\ in source}: share of reference sentences appearing in the source near-verbatim / as an exact substring.}
\label{tab:leakage}
\resizebox{0.6\textwidth}{!}{%
\begin{tabular}{l l l l c}
\toprule
Dataset & Source & Target & Header check & Ref.\ in source \\
\midrule
ACI-Bench & Dialogue & Visit note & n/a & 0.0\% / 1.2\% \\
MIMIC-BHC & Discharge note minus BHC & Brief Hospital Course & 0/500 contain BHC & 0.8\% / 1.0\% \\
MIMIC-CXR & FINDINGS & IMPRESSION & 0/500 contain IMPRESSION & 2.2\% / 6.2\% \\
Priv-DS & Encounter notes & Discharge summary & 0 shared sentences & 22.4\% / 23.6\% \\
\bottomrule
\end{tabular}}
\end{table}

Splits are random at the document level (admission or report). The public derived releases we use do not include subject identifiers, so we searched for likely repeat patients by content: near-identical patient-specific history text, confirmed with two independent similarity methods. We found no likely pairs in ACI-Bench or MIMIC-CXR, 3 pairs in MIMIC-BHC (6/500 documents), and 2 pairs in Priv-DS (4/500). Removing the affected test documents changed test risk by at most 0.1 percentage points, and the change was upward. Because the recall of a content-based search is unknown, especially for short radiology reports, this does not establish patient independence. Patient-grouped splits are left to future work.

\subsection{Copy-Forward in Priv-DS}
\label{app:copy-forward}
Clinical notes often copy text forward across notes. To assess its effect on Priv-DS, we detected within-admission near-duplicates with a pre-specified rule: normalized token-set Jaccard $\ge 0.8$ on sentences of at least five tokens. We checked sensitivity at thresholds from 0.6 to 1.0. In total, 19.9\% of source sentences repeat earlier content in the same admission (16.5\% exact repeats; 16.5--23.9\% across thresholds), and 484 of 500 admissions contain at least one repeat. Duplicate pairs receive the same oracle importance label in 93.6\% of cases, so repetition does not introduce inconsistent labels. At the deployed thresholds, duplicate pairs receive the same surfacing decision in 92.9\% of cases. The per-admission omission miss rate is 9.1\% when counting sentences and 8.7\% when counting distinct content, where a copied omission counts as caught if any copy is surfaced. These rates are in-sample over the 499 admissions with at least one true omission. About 19\% of surfaced sentences repeat content already surfaced earlier in the same admission, so a display that collapses duplicates would reduce review effort without changing the surfaced set or the guarantee.

\section{Detailed Results}
\label{app:detailed}

\paragraph{Calibrated thresholds.}

Table~\ref{tab:thresholds} reports the calibrated thresholds selected at $\alpha=0.15$ for each dataset. The hallucination threshold $\lambda^*$ is applied to the support score, while the omission thresholds $(\tau^*,\gamma^*)$ are applied to importance and non-coverage scores.

\begin{table}[!htbp]
\centering
\small
\setlength{\tabcolsep}{10pt}

\caption{Calibrated thresholds at $\alpha=0.15$ from the single canonical 70/30 calibration split (seed 42). All risk and workload results elsewhere average 100 resplits, each of which recalibrates its own thresholds. Summary sentences are flagged when $\hat p_{\mathrm{sup}} \le \lambda^*$. Source sentences are surfaced when $\hat p_{\mathrm{imp}} \ge \tau^*$ and $\hat p_{\mathrm{ncov}} \ge \gamma^*$, where $\hat p_{\mathrm{ncov}}=1-\hat p_{\mathrm{cov}}$.}
\label{tab:thresholds}
\begin{tabular}{l ccc}
\toprule
Dataset & $\lambda^*$ & $\tau^*$ & $\gamma^*$ \\
\midrule
ACI-Bench      & 0.70 & 0.15 & 0.05 \\
MIMIC-BHC      & 0.60 & 0.30 & 0.05 \\
MIMIC-CXR      & 0.50 & 0.05 & 0.45 \\
Priv-DS        & 0.80 & 0.45 & 0.05 \\
SumPubMed      & 0.50 & 0.30 & 0.10 \\
\bottomrule
\end{tabular}
\end{table}

Two patterns are notable. First, ACI-Bench and Priv-DS select higher hallucination thresholds ($\lambda^*=0.70$--$0.80$), indicating that these domains require more aggressive flagging of lower-support summary sentences to meet the target risk level. Second, the deployed omission policies differ in which gate does the work. On ACI-Bench, MIMIC-BHC, and Priv-DS, the minimum-workload certified cell pairs a moderate importance threshold with $\gamma^*=0.05$. Its coverage gate removes only sentences that all five judge votes deem fully covered, which are the important-but-already-summarized sentences that importance-only selection must still surface. On MIMIC-CXR, the importance gate is nearly off ($\tau^*=0.05$) and the coverage gate carries the selection ($\gamma^*=0.45$), consistent with the limited separability of importance scores in short radiology reports. Which gate dominates is determined by calibration rather than fixed in advance, which a single-score or single-gate controller cannot replicate.

\paragraph{Reproducibility details.}
\label{app:repro}
Table~\ref{tab:repro} lists the settings needed to reproduce all experiments. Code, prompts, configuration files, and per-sentence judge scores are released with the paper.

\begin{table}[!htbp]
\centering
\scriptsize
\setlength{\tabcolsep}{3pt}
\caption{Reproducibility settings.}
\label{tab:repro}
\begin{tabular}{>{\raggedright\arraybackslash}p{2.2cm} >{\raggedright\arraybackslash}p{4.9cm}}
\toprule
Item & Setting \\
\midrule
Summarizer & Llama-3.3-70B-Instruct (Microsoft Azure), temperature 0 \\
Oracle & GPT-5 (\texttt{gpt-5-2025-08-07}), provider-default sampling, two-pass propose-and-verify \\
Judge & GPT-5-mini (\texttt{gpt-5-mini-2025-08-07}), provider-default sampling, $m=5$ independent calls per sentence; three-tier votes mapped to $\{0,0.5,1\}$ and averaged \\
Alt.\ summarizers & Gemini 2.5 Pro; GPT-5; GPT-5-mini \\
Alt.\ judges & Gemini 2.5 Pro; Qwen3.5-4B, local inference \\
External scorers & \texttt{cross-encoder/nli-deberta-v3-base}; \texttt{bert-base-uncased}; \textsc{AlignScore}-base \\
Output length & \texttt{max\_tokens}=8000 for all LLM calls \\
Hallucination grid & $\lambda\in\{0,0.01,\dots,1\}$, scalar CRC \\
Omission grid & $(\tau,\gamma)\in\{0,0.05,\dots,1\}^2$ (441 cells); order $\tau+\gamma$ ascending, ties by $\tau$ then $\gamma$; HB $p$-values; deploy min-workload certified cell \\
Risk levels & $\alpha=0.15$ (main); sweep $\{0.05,0.10,0.15,0.20,0.25,0.30,0.40,0.50\}$; $\delta_{\mathrm{omit}}=0.05$ \\
Splits & Canonical 70/30 split (seed 42) for reported thresholds; 100 random 70/30 document-level resplits (seed 42) for all risk and workload results \\
\bottomrule
\end{tabular}
\end{table}

\section{Calibration Baselines}
\label{app:baselines}

Table~\ref{tab:baselines} compares omission controllers that differ in how they calibrate the importance and non-coverage thresholds. It tests whether two-dimensional calibrated selection reduces review workload while preserving risk control.

\begin{table}[H]
\centering
\footnotesize
\caption{Omission controller comparison at $\alpha=0.15$ (fractional loss), averaged over 100 random 70/30 cal/test resplits. V\% is the mean violation with a descriptive 95\% bootstrap interval over resplit-level estimates. \textbf{First block}: methods with the same high-probability guarantee $\Pr(R\le\alpha)\ge 1-\delta$, $\delta=0.05$ (Learn-Then-Test with Hoeffding--Bentkus $p$-values); \textsc{LTT-FST} is the \textsc{CARE} omission controller. Bold marks fewest flagged sentences in this block. \textbf{Second block}: the same reduced-dimensional baselines calibrated with scalar CRC, which gives the weaker expected-risk guarantee $\mathbb{E}[R]\le\alpha$. \textbf{Third block}: empirical thresholding baselines with no finite-sample guarantee: Max-F1 (oracle F1-optimal $(\tau,\gamma)$ on cal using labels, ignores $\alpha$) and Fixed-0.5 (label-free: $\tau{=}\gamma{=}0.5$, no calibration). Red V\% marks cells whose interval lies entirely above $\alpha$.}
\label{tab:baselines}
\resizebox{0.48\textwidth}{!}{
\begin{tabular}{ll rrr}
\toprule
Dataset & Method & V\% [95\% int.] & Flagged & Rec\% \\
\midrule
ACI-Bench & \textbf{LTT-FST} & \textbf{5.7 [5.0, 6.5]} & \textbf{15.2} (29.3\%) & 88.7 \\
\rowcolor{gray!6}
 & 1D-Imp (HB) & 6.3 [5.3, 7.2] & 29.8 (57.3\%) & 89.4 \\
 & Product (HB) & 6.0 [5.1, 6.9] & 15.9 (30.5\%) & 89.5 \\
\rowcolor{gray!6}
 & Union Bnd (HB) & 0.9 [0.6, 1.3] & 33.6 (64.7\%) & 97.1 \\
\cdashline{2-5}
 & 1D-Imp (CRC) & 11.9 [10.6, 13.2] & 25.2 (48.6\%) & 79.4 \\
\rowcolor{gray!6}
 & Product (CRC) & 14.1 [12.9, 15.4] & 11.0 (21.2\%) & 76.1 \\
 & Union Bnd (CRC) & 11.3 [10.1, 12.5] & 11.5 (22.1\%) & 80.7 \\
\cdashline{2-5}
\rowcolor{gray!6}
 & \textcolor{gray}{Max-F1\textsuperscript{*}} & \textcolor{redflag}{28.1 [27.0, 29.3]} & \textcolor{gray}{4.1 (8.0\%)} & \textcolor{gray}{48.6} \\
 & \textcolor{gray}{Fixed-0.5\textsuperscript{*}} & \textcolor{redflag}{16.1 [15.2, 16.9]} & \textcolor{gray}{8.7 (16.7\%)} & \textcolor{gray}{71.5} \\
\midrule
MIMIC-BHC & \textbf{LTT-FST} & \textbf{10.4 [10.0, 10.7]} & \textbf{71.4} (62.0\%) & 89.5 \\
\rowcolor{gray!6}
 & 1D-Imp (HB) & 10.0 [9.8, 10.1] & 87.7 (76.1\%) & 89.6 \\
 & Product (HB) & 9.8 [9.5, 10.1] & 72.9 (63.3\%) & 90.8 \\
\rowcolor{gray!6}
 & Union Bnd (HB) & 6.8 [6.5, 7.1] & 78.0 (67.7\%) & 93.9 \\
\cdashline{2-5}
 & 1D-Imp (CRC) & 11.9 [11.2, 12.5] & 84.7 (73.5\%) & 87.3 \\
\rowcolor{gray!6}
 & Product (CRC) & 10.1 [9.9, 10.3] & 72.3 (62.8\%) & 90.5 \\
 & Union Bnd (CRC) & 10.1 [9.9, 10.3] & 71.8 (62.4\%) & 90.6 \\
\cdashline{2-5}
\rowcolor{gray!6}
 & \textcolor{gray}{Max-F1\textsuperscript{*}} & \textcolor{redflag}{40.3 [40.1, 40.6]} & \textcolor{gray}{30.0 (26.0\%)} & \textcolor{gray}{58.9} \\
 & \textcolor{gray}{Fixed-0.5\textsuperscript{*}} & \textcolor{redflag}{18.6 [18.4, 18.9]} & \textcolor{gray}{58.6 (50.9\%)} & \textcolor{gray}{80.7} \\
\midrule
MIMIC-CXR & LTT-FST & 2.0 [1.8, 2.2] & 2.1 (26.4\%) & 90.4 \\
\rowcolor{gray!6}
 & 1D-Imp (HB) & 7.6 [7.2, 7.9] & 4.9 (61.8\%) & 67.1 \\
 & Product (HB) & 5.4 [5.1, 5.8] & 1.5 (19.1\%) & 75.1 \\
\rowcolor{gray!6}
 & \textbf{Union Bnd (HB)} & \textbf{5.5 [5.1, 5.9]} & \textbf{1.2} (15.4\%) & 74.9 \\
\cdashline{2-5}
 & 1D-Imp (CRC) & 7.6 [7.2, 7.9] & 4.9 (61.8\%) & 67.1 \\
\rowcolor{gray!6}
 & Product (CRC) & 5.6 [5.2, 6.2] & 1.5 (18.9\%) & 74.5 \\
 & Union Bnd (CRC) & 8.1 [7.6, 8.7] & 1.0 (12.6\%) & 63.7 \\
\cdashline{2-5}
\rowcolor{gray!6}
 & \textcolor{gray}{Max-F1\textsuperscript{*}} & \textcolor{gray}{7.2 [6.7, 7.7]} & \textcolor{gray}{1.1 (13.8\%)} & \textcolor{gray}{67.6} \\
 & \textcolor{gray}{Fixed-0.5\textsuperscript{*}} & \textcolor{gray}{2.3 [2.1, 2.4]} & \textcolor{gray}{1.8 (23.0\%)} & \textcolor{gray}{89.6} \\
\midrule
Priv-DS & \textbf{LTT-FST} & \textbf{9.1 [9.0, 9.2]} & \textbf{235.8} (65.7\%) & 90.9 \\
\rowcolor{gray!6}
 & 1D-Imp (HB) & 8.8 [8.7, 8.9] & 261.6 (72.9\%) & 91.2 \\
 & Product (HB) & 7.4 [7.4, 7.5] & 250.0 (69.7\%) & 92.8 \\
\rowcolor{gray!6}
 & Union Bnd (HB) & 6.8 [6.6, 6.9] & 255.8 (71.3\%) & 93.5 \\
\cdashline{2-5}
 & 1D-Imp (CRC) & 8.8 [8.7, 8.9] & 261.6 (72.9\%) & 91.2 \\
\rowcolor{gray!6}
 & Product (CRC) & 13.0 [12.8, 13.2] & 228.4 (63.6\%) & 87.4 \\
 & Union Bnd (CRC) & 9.2 [9.1, 9.3] & 234.6 (65.4\%) & 91.1 \\
\cdashline{2-5}
\rowcolor{gray!6}
 & \textcolor{gray}{Max-F1\textsuperscript{*}} & \textcolor{redflag}{27.4 [27.3, 27.6]} & \textcolor{gray}{151.1 (42.1\%)} & \textcolor{gray}{72.3} \\
 & \textcolor{gray}{Fixed-0.5\textsuperscript{*}} & \textcolor{gray}{11.7 [11.6, 11.8]} & \textcolor{gray}{219.0 (61.0\%)} & \textcolor{gray}{88.6} \\
\midrule
SumPubMed & LTT-FST & 7.4 [7.3, 7.6] & 122.4 (72.2\%) & 92.1 \\
\rowcolor{gray!6}
 & 1D-Imp (HB) & 6.9 [6.8, 7.0] & 137.8 (81.3\%) & 92.5 \\
 & \textbf{Product (HB)} & \textbf{10.1 [10.0, 10.3]} & \textbf{121.0} (71.4\%) & 89.8 \\
\rowcolor{gray!6}
 & Union Bnd (HB) & 7.1 [7.0, 7.3] & 127.2 (75.1\%) & 92.7 \\
\cdashline{2-5}
 & 1D-Imp (CRC) & 6.9 [6.8, 7.0] & 137.8 (81.3\%) & 92.5 \\
\rowcolor{gray!6}
 & Product (CRC) & 10.1 [10.0, 10.3] & 121.0 (71.4\%) & 89.8 \\
 & Union Bnd (CRC) & 11.9 [11.7, 12.1] & 113.2 (66.8\%) & 87.8 \\
\cdashline{2-5}
\rowcolor{gray!6}
 & \textcolor{gray}{Max-F1\textsuperscript{*}} & \textcolor{redflag}{47.8 [47.5, 48.1]} & \textcolor{gray}{38.4 (22.7\%)} & \textcolor{gray}{52.4} \\
 & \textcolor{gray}{Fixed-0.5\textsuperscript{*}} & \textcolor{redflag}{15.2 [15.1, 15.4]} & \textcolor{gray}{109.0 (64.3\%)} & \textcolor{gray}{84.5} \\
\bottomrule
\multicolumn{5}{l}{\textsuperscript{*}No finite-sample guarantee; empirical numbers shown for comparison.} \\
\end{tabular}}
\end{table}

\textbf{LTT-FST}, the \textsc{CARE} omission controller, applies fixed-sequence testing to the pre-specified two-dimensional grid. \textbf{1D-Imp} calibrates a single importance threshold $\tau$ and disables the non-coverage gate by setting $\gamma=0$. This adapts conformal importance selection \citep{kuwahara2025} from extractive summarization to our omission loss. \textbf{Product} forms the scalar composite score
\[
s_{\mathrm{prod}}(u)
=
\hat p_{\mathrm{imp}}(u)
\hat p_{\mathrm{ncov}}(u)
\]
and calibrates a single threshold $\beta$ on this score. \textbf{Union Bound} calibrates the importance and non-coverage gates independently and surfaces their intersection. A true omission missed by the intersection must be missed by at least one marginal gate, so the joint omission loss is bounded by the sum of the two marginal miss losses.

Each reduced-dimensional baseline is reported under two calibration procedures. The \emph{HB} variants carry the same high-probability guarantee as LTT-FST, $\Pr(R\le\alpha)\ge1-\delta$. 1D-Imp and Product run fixed-sequence HB testing along their scalar threshold, and Union splits both budgets as $(\alpha/2,\delta/2)$ across the two gates. The \emph{CRC} variants use scalar CRC, which gives the weaker expected-risk guarantee $\mathbb E[R]\le\alpha$. For Union, the CRC variant uses budgets $\alpha/2+\alpha/2$. Comparisons are like-for-like only within a block. Below the second dashed line are two empirical thresholding baselines without finite-sample guarantees. \textbf{Max-F1} selects $(\tau,\gamma)$ to maximize calibration-set F1 without a risk constraint, and \textbf{Fixed-0.5} applies the label-free heuristic $\tau{=}\gamma{=}0.5$ with no calibration.

For scalar omission baselines, increasing the displayed threshold reduces the surfaced set and increases omission loss. We therefore calibrate with the parameter ordering reversed, equivalently using permissiveness variables such as $1-\tau$ and $1-\beta$.

\paragraph{An active second gate must be included in calibration.}
The Partial baseline first calibrates the importance threshold and then applies an additional uncalibrated non-coverage gate. This post-hoc gate removes surfaced sentences and can invalidate the original one-dimensional guarantee: fractional violation reaches 49.8\% (Table~\ref{tab:binary-vs-fractional}). In contrast, 1D-Imp remains valid because it disables the second gate entirely by setting $\gamma=0$. Thus, whenever both importance and non-coverage affect the final surfacing decision, their joint effect must be accounted for during calibration.

\paragraph{Workload at a common guarantee.}
Within the HB block, all methods control risk, and LTT-FST surfaces the fewest sentences on ACI-Bench, MIMIC-BHC, and Priv-DS. Union-HB and Product-HB are lower on MIMIC-CXR, and Product-HB is marginally lower on SumPubMed. 1D-Imp surfaces $1.1$--$2.3\times$ more source sentences than LTT-FST because, without a coverage signal, every important sentence must be reviewed, including those the summary already covers. Union-HB is conservative on long documents because it splits both $\alpha$ and $\delta$. Realized violation differs across methods, so these comparisons should not be read as Pareto dominance at matched realized risk. The CRC block operates closer to $\alpha$ and often at lower workload than the HB block because its guarantee is in expectation rather than with high probability. The uncalibrated baselines sometimes reduce workload further but frequently exceed $\alpha$ (Max-F1 on four of five datasets).

\section{Scorer Agnosticism}
\label{app:scorer-agnostic}

To test whether \textsc{CARE}'s risk control depends on the specific LLM judge, we replace the support and coverage scorers with three external scorer families: (i) a DeBERTa-v3 NLI cross-encoder, using reverse-direction NLI for coverage, (ii) BERT-base CLS-token cosine similarity, using max-similarity for both controllers, and (iii) \textsc{AlignScore} \citep{zha-alignscore-2023}, a unified alignment classifier. We hold the importance scorer fixed to the LLM-judge signal, since no standard reference-free salience scorer exists. This experiment therefore tests whether \textsc{CARE} remains valid under variation in the support and coverage signals most commonly used in faithfulness evaluation.

For each dataset--scorer pair, we rerun 100 random 70/30 calibration/test splits and compare \textsc{CARE} with Max-F1 thresholding, which selects thresholds to maximize calibration-set F1 without enforcing a risk constraint.

\begin{table}[!t]
\centering
\scriptsize
\setlength{\tabcolsep}{4.5pt}
\caption{Scorer-agnostic risk control at $\alpha=0.15$ over 100 random 70/30 calibration/test resplits. The LLM-judge support and coverage signals are replaced with NLI, embedding cosine, or \textsc{AlignScore} scorers, while importance is held fixed to the LLM judge. Each cell reports mean violation percentage $\pm$ half-width of a descriptive 95\% normal-approximation interval over resplit-level estimates. $\dagger$ marks Max-F1 cells whose interval lies entirely above $\alpha$.}
\label{tab:external-scorers}
\resizebox{0.6\textwidth}{!}{%
\begin{tabular}{ll rr rr}
\toprule
 & & \multicolumn{2}{c}{Hallucination V\%} & \multicolumn{2}{c}{Omission V\%} \\
\cmidrule(lr){3-4}\cmidrule(lr){5-6}
Dataset & Scorer & \textsc{CARE} & Max-F1 & \textsc{CARE} & Max-F1 \\
\midrule
ACI-Bench & NLI (DeBERTa) & 15.9{\scriptsize $\pm$1.4} & \textbf{28.9}$^\dagger${\scriptsize $\pm$1.4} & 6.4{\scriptsize $\pm$0.9} & \textbf{25.0}$^\dagger${\scriptsize $\pm$1.3} \\
 & Embed (BERT) & 11.2{\scriptsize $\pm$1.4} & \textbf{26.1}$^\dagger${\scriptsize $\pm$4.1} & 3.3{\scriptsize $\pm$0.4} & \textbf{28.7}$^\dagger${\scriptsize $\pm$1.5} \\
 & AlignScore & 14.8{\scriptsize $\pm$1.3} & \textbf{49.5}$^\dagger${\scriptsize $\pm$1.9} & 6.2{\scriptsize $\pm$1.0} & \textbf{36.5}$^\dagger${\scriptsize $\pm$1.3} \\
\midrule
MIMIC-BHC & NLI (DeBERTa) & 15.2{\scriptsize $\pm$0.6} & \textbf{18.5}$^\dagger${\scriptsize $\pm$1.7} & 10.2{\scriptsize $\pm$0.3} & \textbf{39.3}$^\dagger${\scriptsize $\pm$0.3} \\
 & Embed (BERT) & 11.0{\scriptsize $\pm$0.4} & 11.9{\scriptsize $\pm$2.4} & 2.2{\scriptsize $\pm$0.1} & \textbf{37.0}$^\dagger${\scriptsize $\pm$0.3} \\
 & AlignScore & 15.0{\scriptsize $\pm$0.7} & \textbf{39.5}$^\dagger${\scriptsize $\pm$0.9} & 10.0{\scriptsize $\pm$0.2} & \textbf{43.4}$^\dagger${\scriptsize $\pm$0.6} \\
\midrule
MIMIC-CXR & NLI (DeBERTa) & 13.6{\scriptsize $\pm$0.7} & 9.5{\scriptsize $\pm$1.0} & 10.0{\scriptsize $\pm$0.5} & 9.7{\scriptsize $\pm$0.5} \\
 & Embed (BERT) & 13.6{\scriptsize $\pm$0.6} & 15.4{\scriptsize $\pm$0.7} & 7.4{\scriptsize $\pm$0.3} & 9.7{\scriptsize $\pm$1.4} \\
 & AlignScore & 15.0{\scriptsize $\pm$0.7} & 13.2{\scriptsize $\pm$0.8} & 10.3{\scriptsize $\pm$0.5} & \textbf{16.4}$^\dagger${\scriptsize $\pm$0.7} \\
\midrule
Priv-DS & NLI (DeBERTa) & 13.1{\scriptsize $\pm$0.4} & \textbf{67.1}$^\dagger${\scriptsize $\pm$0.6} & 9.6{\scriptsize $\pm$0.1} & \textbf{26.0}$^\dagger${\scriptsize $\pm$0.2} \\
 & Embed (BERT) & 11.6{\scriptsize $\pm$0.4} & 4.1{\scriptsize $\pm$0.3} & 2.7{\scriptsize $\pm$0.0} & \textbf{25.0}$^\dagger${\scriptsize $\pm$0.2} \\
 & AlignScore & 13.8{\scriptsize $\pm$0.6} & \textbf{71.3}$^\dagger${\scriptsize $\pm$0.7} & 9.0{\scriptsize $\pm$0.1} & \textbf{27.6}$^\dagger${\scriptsize $\pm$0.2} \\
\midrule
SumPubMed & NLI (DeBERTa) & 9.7{\scriptsize $\pm$0.4} & 9.7{\scriptsize $\pm$0.4} & 6.9{\scriptsize $\pm$0.1} & \textbf{39.3}$^\dagger${\scriptsize $\pm$0.5} \\
 & Embed (BERT) & 12.9{\scriptsize $\pm$0.5} & \textbf{26.5}$^\dagger${\scriptsize $\pm$2.2} & 2.8{\scriptsize $\pm$0.1} & \textbf{38.8}$^\dagger${\scriptsize $\pm$0.5} \\
 & AlignScore & 15.1{\scriptsize $\pm$0.7} & \textbf{27.5}$^\dagger${\scriptsize $\pm$0.8} & 7.4{\scriptsize $\pm$0.1} & \textbf{43.4}$^\dagger${\scriptsize $\pm$0.6} \\
\bottomrule
\end{tabular}}
\end{table}

Table~\ref{tab:external-scorers} shows that \textsc{CARE} controls risk across external scorer families. Omission violation is 2.2--10.3\% in all 15 dataset--scorer settings, below $\alpha$ in every case. Hallucination violation is 9.7--15.9\%; four means lie slightly above $\alpha$ (NLI on ACI-Bench, 15.9\%, and on MIMIC-BHC, 15.2\%; \textsc{AlignScore} on MIMIC-CXR and SumPubMed, 15.0\% and 15.1\%), by at most 0.9 points and with intervals that include $\alpha$. CRC bounds the risk in expectation over calibration draws, so overshoots of this size are consistent with finite-sample variation. The resplit intervals summarize sensitivity to the calibration/test partition of the fixed document pool and should not be interpreted as population confidence intervals or as evidence that these overshoots disappear under repeated sampling. In contrast, Max-F1 thresholding exceeds the target, with its interval entirely above $\alpha$, in 22 of 30 dataset--scorer--controller settings (9 of 15 for hallucination, 13 of 15 for omission), because it optimizes calibration-set F1 without controlling missed-error risk. Overall, these results support the narrower claim that the calibration layer can be applied to different scorer families, while the realized safety--efficiency tradeoff depends strongly on the scorer.

As a complementary check within the LLM-judge family, we also replace the primary Phase~2 judge with Gemini 2.5 Pro and Qwen3.5-4B on ACI-Bench and MIMIC-BHC. Across 100 resplits, \textsc{CARE} maintains violation rates near or below $\alpha$ for both controllers and both alternative judges. With Gemini 2.5 Pro, violation rates are 13.4\% hallucination / 6.6\% omission on ACI-Bench and 10.0\% / 10.4\% on MIMIC-BHC. With Qwen3.5-4B, violation rates are 0.0\% hallucination / 4.4\% omission on ACI-Bench and 0.0\% / 9.8\% on MIMIC-BHC. Both judges are recalibrated with the same CRC and LTT-FST procedures over the same 100 resplits (\texttt{compute\_fst\_alt\_judges.py}).

\paragraph{Validity does not imply efficiency.}
Qwen3.5-4B illustrates this distinction clearly. Its calibrated hallucination controller attains zero held-out violation by flagging 100\% of summary sentences on both evaluated datasets. Scorer agnosticism therefore refers to the ability of calibration to control risk. A weak or poorly aligned scorer can still yield an impractically conservative operating point.

\paragraph{Efficiency is not scorer-agnostic.} Although risk control is largely scorer-agnostic after calibration, efficiency is not. Table~\ref{tab:external-scorers-efficiency} compares per-document workload and recall for the LLM judge against external scorers at the same $\alpha=0.15$. For hallucination control, the LLM judge reaches the target with fewer flags at comparable or higher recall. On MIMIC-CXR, external scorers retain only 53--60\% hallucination recall, compared with 83\% for the LLM judge. The same holds for omission control: GPT-5-mini surfaces the fewest source sentences on every dataset (26.4--72.2\%, against 31.9--90.7\% for external scorers), and on MIMIC-CXR external scorers reach only 56--66\% omission recall, compared with 90\% for the LLM judge. Because the primary oracle and judge are both from the GPT model family, these experiments do not isolate whether GPT-5-mini's efficiency advantage reflects general scorer quality or closer alignment with the oracle labeling function.

\begin{table}[!htbp]
\centering
\scriptsize
\setlength{\tabcolsep}{3pt}
\caption{Scorer efficiency at $\alpha=0.15$ (full version of Table~\ref{tab:scorers-compact}). All scorers use \textsc{CARE} calibration over the same 100 random 70/30 calibration/test resplits; importance is held fixed to the LLM judge for external scorers. \emph{Flagged}: mean sentences flagged or surfaced per document (\% of total). \emph{Rec\%}: mean recall of oracle-labeled hallucinations or omissions. Validity for each scorer is reported in Table~\ref{tab:external-scorers}.}
\label{tab:external-scorers-efficiency}
\resizebox{0.55\textwidth}{!}{%
\begin{tabular}{lll rr rr}
\toprule
 & & & \multicolumn{2}{c}{Hallucination} & \multicolumn{2}{c}{Omission} \\
\cmidrule(lr){4-5}\cmidrule(lr){6-7}
Dataset & Type & Scorer & Flagged & Rec\% & Flagged & Rec\% \\
\midrule
  \multirow{6}{*}{\textbf{ACI-Bench}} & \multirow{3}{*}{LLM} & \textbf{GPT-5-mini} & \textbf{3.7} (13.7\%) & \textbf{87.5} & \textbf{15.2} (29.3\%) & \textbf{88.7} \\
   &  & Gemini 2.5 Pro & 5.0 (17.4\%) & 88.4 & 18.6 (35.4\%) & 87.0 \\
   &  & Qwen3.5-4B & 28.4 (100\%) & 100.0 & 43.0 (83.5\%) & 93.9 \\
\addlinespace[2pt]
   & \multirow{3}{*}{External} & NLI (DeBERTa) & 13.0 (48.3\%) & 86.9 & 29.1 (56.6\%) & 89.2 \\
   &  & Embed (BERT) & 23.2 (86.0\%) & 89.6 & 33.3 (64.8\%) & 94.6 \\
   &  & AlignScore & 17.8 (66.3\%) & 85.9 & 29.2 (56.8\%) & 89.6 \\
\midrule
  \multirow{6}{*}{\textbf{MIMIC-BHC}} & \multirow{3}{*}{LLM} & \textbf{GPT-5-mini} & \textbf{2.3} (13.5\%) & \textbf{85.7} & \textbf{71.4} (62.0\%) & \textbf{89.5} \\
   &  & Gemini 2.5 Pro & 2.5 (14.5\%) & 88.7 & 73.6 (63.8\%) & 90.4 \\
   &  & Qwen3.5-4B & 17.1 (100\%) & 100.0 & 85.4 (74.0\%) & 90.3 \\
\addlinespace[2pt]
   & \multirow{3}{*}{External} & NLI (DeBERTa) & 10.4 (61.2\%) & 81.2 & 87.4 (75.8\%) & 89.5 \\
   &  & Embed (BERT) & 14.5 (84.8\%) & 87.1 & 103.1 (89.4\%) & 98.0 \\
   &  & AlignScore & 10.0 (58.7\%) & 82.2 & 83.8 (72.7\%) & 89.6 \\
\midrule
  \multirow{4}{*}{\textbf{MIMIC-CXR}} & LLM & \textbf{GPT-5-mini} & \textbf{1.2} (24.4\%) & \textbf{82.9} & \textbf{2.1} (26.4\%) & \textbf{90.4} \\
\addlinespace[2pt]
   & \multirow{3}{*}{External} & NLI (DeBERTa) & 2.1 (40.2\%) & 58.4 & 2.9 (36.8\%) & 56.2 \\
   &  & Embed (BERT) & 2.2 (44.1\%) & 59.5 & 5.1 (64.0\%) & 66.1 \\
   &  & AlignScore & 1.8 (34.9\%) & 52.9 & 2.5 (31.9\%) & 55.9 \\
\midrule
  \multirow{4}{*}{\textbf{Priv-DS}} & LLM & \textbf{GPT-5-mini} & \textbf{8.1} (22.6\%) & \textbf{94.9} & \textbf{235.8} (65.7\%) & \textbf{90.9} \\
\addlinespace[2pt]
   & \multirow{3}{*}{External} & NLI (DeBERTa) & 26.6 (74.2\%) & 95.1 & 257.2 (71.8\%) & 90.5 \\
   &  & Embed (BERT) & 34.0 (94.9\%) & 95.0 & 309.0 (86.3\%) & 97.4 \\
   &  & AlignScore & 27.5 (76.8\%) & 95.0 & 256.5 (71.6\%) & 91.1 \\
\midrule
  \multirow{4}{*}{\textbf{SumPubMed}} & LLM & \textbf{GPT-5-mini} & \textbf{1.0} (7.7\%) & \textbf{86.2} & \textbf{122.4} (72.2\%) & \textbf{92.1} \\
\addlinespace[2pt]
   & \multirow{3}{*}{External} & NLI (DeBERTa) & 6.5 (49.3\%) & 76.2 & 137.1 (81.1\%) & 92.4 \\
   &  & Embed (BERT) & 7.2 (55.1\%) & 71.3 & 153.4 (90.7\%) & 96.8 \\
   &  & AlignScore & 2.6 (20.2\%) & 65.4 & 134.2 (79.3\%) & 92.1 \\
\bottomrule
\end{tabular}}
\end{table}

\paragraph{Detectors at their default threshold.}
Off-the-shelf detectors are often deployed at a fixed decision threshold of 0.5 with no calibration. Table~\ref{tab:default-threshold} evaluates this practice on the document-level hallucination loss over the same 100 resplits. The flagging rule and settings were pre-registered before any run. BERT cosine is excluded because its scores are not probabilities, and MIMIC-BHC external scorers were not run because of compute limits. The default threshold misses $\alpha$ in 5 of 7 settings. Where it meets the target, NLI on MIMIC-CXR does so by flagging 58\% of summary sentences. Calibration moves the threshold in whichever direction is needed: it raises flagging where the default under-flags and lowers it where the default over-flags. The reliability of the default cannot be anticipated without labels.

% \begin{table}[!htbp]
% \centering
% \scriptsize
% \setlength{\tabcolsep}{3pt}
% \caption{Document-level hallucination violation at the default 0.5 threshold vs.\ after \textsc{CARE} calibration ($\alpha=0.15$, 100 resplits). \emph{Flagged}: share of summary sentences flagged, default $\to$ calibrated. LLM judge at default uses majority vote. $^\dagger$Exceeds $\alpha$.}
% \label{tab:default-threshold}
% \begin{tabular}{ll cc c}
% \toprule
% Dataset & Scorer & Default 0.5 & \textsc{CARE} & Flagged \\
% \midrule
% ACI-Bench & LLM judge & 18.3$^\dagger$ & 12.2 & 11.7 $\to$ 13.7\% \\
%  & NLI (DeBERTa) & 18.8$^\dagger$ & 15.9 & 43 $\to$ 48\% \\
%  & \textsc{AlignScore} & 19.6$^\dagger$ & 14.8 & 56 $\to$ 66\% \\
% MIMIC-BHC & LLM judge & 16.7$^\dagger$ & 12.5 & 11.0 $\to$ 13.5\% \\
% MIMIC-CXR & LLM judge & 5.7 & 5.7 & 24 $\to$ 24\% \\
%  & NLI (DeBERTa) & 6.1 & 13.6 & 58 $\to$ 40\% \\
%  & \textsc{AlignScore} & 15.8$^\dagger$ & 15.0 & 33 $\to$ 35\% \\
% \bottomrule
% \end{tabular}
% \end{table}

\begin{table}[!htbp]
\centering
\scriptsize
\setlength{\tabcolsep}{3pt}
\caption{Document-level hallucination violation at the default 0.5 threshold vs.\ after \textsc{CARE} calibration ($\alpha=0.15$, 100 resplits). \emph{Flagged}: share of summary sentences flagged, default $\to$ calibrated. LLM judge at default uses majority vote. $^\dagger$Exceeds $\alpha$.}
\label{tab:default-threshold}
\resizebox{0.55\textwidth}{!}{%
\begin{tabular}{@{}llccc@{}}
\toprule
Dataset & Scorer & Default 0.5 & \textsc{CARE} & Flagged \\
\midrule
ACI-Bench & LLM judge & 18.3$^\dagger$ & 12.2 & 11.7 $\to$ 13.7\% \\
 & NLI (DeBERTa) & 18.8$^\dagger$ & 15.9 & 43 $\to$ 48\% \\
 & \textsc{AlignScore} & 19.6$^\dagger$ & 14.8 & 56 $\to$ 66\% \\
MIMIC-BHC & LLM judge & 16.7$^\dagger$ & 12.5 & 11.0 $\to$ 13.5\% \\
MIMIC-CXR & LLM judge & 5.7 & 5.7 & 24 $\to$ 24\% \\
 & NLI (DeBERTa) & 6.1 & 13.6 & 58 $\to$ 40\% \\
 & \textsc{AlignScore} & 15.8$^\dagger$ & 15.0 & 33 $\to$ 35\% \\
\bottomrule
\end{tabular}%
}
\end{table}

\section{Prompt Sensitivity Ablation}
\label{app:prompt-ablation}

We compare generic judge prompts with domain-specific judge prompts across all five datasets. Generic prompts use the same tier definitions across datasets, while domain-specific prompts replace the factuality and importance criteria with task-specific guidance. The coverage prompt is held fixed across both conditions. For each prompt condition, we rerun 100 random 70/30 calibration/test resplits.

Table~\ref{tab:prompt-ablation} shows that violation remains controlled under both prompt styles. Prompt choice affects workload in a domain-dependent way: domain-specific prompts reduce omission flags on MIMIC-CXR (26.4\%$\to$21.7\%) and SumPubMed (72.2\%$\to$65.4\%), increase them on ACI-Bench (29.3\%$\to$33.0\%), and have little effect on MIMIC-BHC and Priv-DS. These results suggest that calibration absorbs many prompt-quality differences, while prompt design can still affect efficiency.

\begin{table}[!htbp]
\centering
\footnotesize
\setlength{\tabcolsep}{2.8pt}
\renewcommand{\arraystretch}{0.82}
\caption{Prompt ablation at $\alpha=0.15$, averaged over 100 random 70/30 calibration/test resplits. \emph{V\%}: violation rate with a descriptive 95\% bootstrap interval over resplit-level estimates. \emph{Fl\%}: flagged or surfaced sentences as a percentage of total sentences.}
\label{tab:prompt-ablation}
\resizebox{0.48\textwidth}{!}{%
\begin{tabular}{ll rr rr}
\toprule
& & \multicolumn{2}{c}{Hallucination} & \multicolumn{2}{c}{Omission} \\
\cmidrule(lr){3-4} \cmidrule(lr){5-6}
Dataset & Prompt & V\% & Fl\% & V\% & Fl\% \\
\midrule
ACI-Bench & Generic & 12.2 {\tiny [10.9,13.4]} & 13.7 & 5.7 {\tiny [5.0,6.5]} & 29.3 \\
 & Domain & 13.2 {\tiny [11.9,14.5]} & 11.2 & 6.3 {\tiny [5.7,7.0]} & 33.0 \\
\midrule
MIMIC-BHC & Generic & 12.5 {\tiny [12.1,13.0]} & 13.5 & 10.4 {\tiny [10.0,10.7]} & 62.0 \\
 & Domain & 12.8 {\tiny [12.2,13.3]} & 11.3 & 10.5 {\tiny [10.3,10.8]} & 62.9 \\
\midrule
MIMIC-CXR & Generic & 5.7 {\tiny [5.4,6.0]} & 24.4 & 2.0 {\tiny [1.8,2.2]} & 26.4 \\
 & Domain & 7.6 {\tiny [7.2,8.0]} & 20.9 & 2.6 {\tiny [2.4,2.8]} & 21.7 \\
\midrule
Priv-DS & Generic & 12.8 {\tiny [12.1,13.5]} & 22.6 & 9.1 {\tiny [9.0,9.2]} & 65.7 \\
 & Domain & 12.7 {\tiny [12.3,13.2]} & 21.9 & 9.2 {\tiny [8.9,9.6]} & 65.8 \\
\midrule
SumPubMed & Generic & 6.8 {\tiny [6.5,7.1]} & 7.7 & 7.4 {\tiny [7.3,7.6]} & 72.2 \\
 & Domain & 13.3 {\tiny [12.6,14.0]} & 7.5 & 10.3 {\tiny [10.1,10.4]} & 65.4 \\
\bottomrule
\end{tabular}}
\renewcommand{\arraystretch}{1}
\end{table}

\paragraph{Generic prompts.}
The generic condition uses task-agnostic rubric definitions. Only the role and document type are filled in from a domain specification (e.g., role = ``clinician'', source = ``patient-doctor dialogue'', output = ``clinical summary'' for ACI-Bench).

\begin{quote}\small
\textbf{Factuality.}
\texttt{You are a \{role\} verifying factual accuracy. For each sentence, respond with one of: SUPPORTED (clearly supported by the source), PARTIAL (ambiguous, partially supported, or reasonable inference), UNSUPPORTED (contradicts or fabricates details not in source).}
\end{quote}

\begin{quote}\small
\textbf{Importance.}
\texttt{You are a \{role\} deciding what to include in a \{output\_type\}. For each \{source\_type\} sentence, respond with one of: ESSENTIAL (contains information that belongs in a \{output\_type\}), RELEVANT (provides supporting context but would not typically appear in a \{output\_type\}), NOT\_RELEVANT (conversational, procedural, or not relevant to a \{output\_type\}).}
\end{quote}

\begin{quote}\small
\textbf{Coverage.}
\texttt{You are an auditor checking for information loss. For each \{source\_type\} sentence, determine whether its content is represented in the \{output\_type\}. COVERED: the sentence's key information is represented, possibly through paraphrase. PARTIAL: some but not all of the key information is represented. OMITTED: the key information is absent, too vague, or its meaning has been lost.}
\end{quote}

\paragraph{Domain-specific prompts.}
The domain condition replaces the one-line tier definitions with detailed, domain-aware criteria. Coverage uses the same prompt in both conditions. Below we list the factuality and importance tier definitions for each dataset.

\subparagraph{ACI-Bench (clinical dialogue).}

\begin{quote}\small
\textbf{Factuality.}
\texttt{SUPPORTED: The sentence is consistent with the dialogue in overall meaning. Paraphrasing, synthesis, clinical interpretation, and standard documentation conventions are all acceptable. PARTIAL: The sentence refers to something discussed but represents it imprecisely beyond what standard clinical inference supports. UNSUPPORTED: Clear invention or contradiction relative to the dialogue, especially: new meds/doses, new diagnoses, new tests/results, new numeric values, invented timelines, incorrect negations, or plan items not discussed.}
\end{quote}

\begin{quote}\small
\textbf{Importance.}
\texttt{ESSENTIAL: Contains a diagnosis, key symptom/history, test result, medication, or a concrete plan/action that directly affects the assessment or plan. RELEVANT: Provides supporting clinical context but could be safely omitted without changing the resulting assessment and plan. NOT\_RELEVANT: Conversational filler, repetition, logistics, administrative content, or clinical information that does not affect the assessment or plan.}
\end{quote}

\subparagraph{MIMIC-BHC (hospital course).}

\begin{quote}\small
\textbf{Factuality.}
\texttt{SUPPORTED: Faithful to the patient record in clinical meaning. Paraphrasing and compression are acceptable as long as no clinical facts are introduced or distorted. PARTIAL: Core claim is grounded but a clinically relevant detail is imprecise or incomplete, without distorting clinical meaning. UNSUPPORTED: Introduces, contradicts, or misrepresents the record, including: fabricated facts, wrong laterality/entity/test, inverted negations, trend errors (saying a value rose when it fell), omitting a key inflection point that changes clinical meaning, or numbers that contradict the sentence's own narrative.}
\end{quote}

\begin{quote}\small
\textbf{Importance.}
\texttt{ESSENTIAL: Core clinical information, including admission reason, key findings that drove care, final diagnosis, major treatments/med changes, and discharge plan/disposition. RELEVANT: Clinically meaningful context that supports the narrative but could be safely omitted without changing the overall story. NOT\_RELEVANT: Does not belong in a brief narrative: boilerplate, administrative text, repeated information, or content with no patient-specific meaning.}
\end{quote}

\subparagraph{MIMIC-CXR (radiology).}

\begin{quote}\small
\textbf{Factuality.}
\texttt{SUPPORTED: The sentence is consistent with the FINDINGS in meaning, without adding new clinical implications. A radiologist reading the sentence would not learn anything beyond what is already in the FINDINGS. Paraphrasing, synthesis, restating uncertainty with the same strength, and standard radiology boilerplate language are acceptable. PARTIAL: The sentence is clearly anchored to a real finding but adds a clinically meaningful interpretation or modification that is not explicitly stated. UNSUPPORTED: The sentence contains a claim that is not grounded in the FINDINGS, introduces new diagnoses, etiologies, or recommendations, contradicts a stated finding, or substitutes language that changes clinical meaning (e.g., ``unchanged'' for ``intact'').}
\end{quote}

\begin{quote}\small
\textbf{Importance.}
\texttt{ESSENTIAL: Drives the overall impression or changes the clinical interpretation (e.g., acute abnormality, actionable finding, device position that matters). RELEVANT: True and clinically meaningful, but does not change the impression (e.g., stable or chronic findings, detailed negatives, supporting context). NOT\_RELEVANT: Acquisition details, boilerplate, or purely descriptive text. Note: Normal or negative findings are ESSENTIAL only if they are the main conclusion (e.g., ``no acute cardiopulmonary process''). Stable or chronic findings are usually RELEVANT, not ESSENTIAL.}
\end{quote}

\subparagraph{SumPubMed (scientific articles).}

\begin{quote}\small
\textbf{Factuality.}
\texttt{SUPPORTED: Consistent with the article in meaning and certainty. Paraphrasing and compression are OK if it does not add new claims. PARTIAL: Grounded but meaningfully imprecise, with scope or certainty is softened/broadened without introducing wrong facts. UNSUPPORTED: Adds claims not in the article or contradicts it (including invented numbers/statistics, methods, or conclusions).}
\end{quote}

\begin{quote}\small
\textbf{Importance.}
\texttt{ESSENTIAL: Core structured-abstract content: the study aim/question, what was done, the main result(s), and the main conclusion. RELEVANT: Helpful supporting context or secondary detail that can be shortened or omitted without changing the main story. NOT\_RELEVANT: Background context, extended mechanistic detail, granular numbers, or discussion-level speculation.}
\end{quote}

\subparagraph{Priv-DS (multi-note discharge).}

\begin{quote}\small
\textbf{Factuality.}
\texttt{SUPPORTED: Faithful to the patient record in clinical meaning and timing. Paraphrasing and synthesis across notes are allowed if no new facts or stronger certainty are introduced. PARTIAL: Anchored to the record but meaningfully imprecise (e.g., blurred timing, softened/strengthened certainty, slightly wrong specificity) without clear fabrication. UNSUPPORTED: Not grounded in the record or contradicts it. Includes invented diagnoses, procedures, meds, results, wrong negations, wrong timing, or wrong clinical trends.}
\end{quote}

\begin{quote}\small
\textbf{Importance.}
\texttt{ESSENTIAL: Information required to understand the hospitalization: why the patient was admitted, what was found, what was done, the final diagnoses, major treatment decisions, and the discharge plan. If removing the sentence would make the hospitalization story incomplete or misleading, it is ESSENTIAL. RELEVANT: Clinically meaningful context that supports the story but could be omitted without changing the main narrative. NOT\_RELEVANT: Does not belong in a discharge summary narrative, including boilerplate, repetition, routine data, or information that did not affect diagnosis, management, or disposition.}
\end{quote}

\section{Judge Self-Consistency}
\label{app:judge-variance}

The judge (GPT-5-mini) is queried $m=5$ times per sentence using a three-tier rubric mapped to $\{0,0.5,1\}$. We average these votes to obtain the scores used for calibration. To assess the stability of these vote-averaged scores, we measure agreement across stochastic replicates on all five datasets.

\begin{table}[!htbp]
  \centering
  \scriptsize
  \setlength{\tabcolsep}{3pt}
  \renewcommand{\arraystretch}{0.82}
  \caption{LLM judge self-consistency across $m=5$ stochastic vote replicates. We report Fleiss' $\kappa$ and unanimity rate, defined as the percentage of sentences for which all five votes agree.}
  \label{tab:judge-variance}
  \resizebox{0.48\textwidth}{!}{%
  \begin{tabular}{l cc cc cc}
  \toprule
   & \multicolumn{2}{c}{Hallucination} & \multicolumn{2}{c}{Importance} & \multicolumn{2}{c}{Coverage} \\
  \cmidrule(lr){2-3} \cmidrule(lr){4-5} \cmidrule(lr){6-7}
  Dataset & $\kappa$ & Unan.\% & $\kappa$ & Unan.\% & $\kappa$ & Unan.\% \\
  \midrule
    ACI-Bench & 0.82 & 90 & 0.78 & 73 & 0.78 & 74 \\
    MIMIC-BHC & 0.76 & 87 & 0.71 & 63 & 0.85 & 83 \\
    MIMIC-CXR & 0.82 & 84 & 0.74 & 77 & 0.90 & 90 \\
    Priv-DS & 0.79 & 86 & 0.74 & 66 & 0.82 & 81 \\
    SumPubMed & 0.77 & 90 & 0.75 & 69 & 0.78 & 75 \\
  \bottomrule
  \end{tabular}}
  \renewcommand{\arraystretch}{1}
  \end{table}

Table~\ref{tab:judge-variance} shows that the judge is self-consistent across datasets and annotation tasks. Fleiss' $\kappa$ ranges from 0.71 to 0.90, corresponding to substantial to almost-perfect agreement, and 63--90\% of sentences receive unanimous votes. Factuality and coverage are generally more stable than importance, consistent with the greater subjectivity of deciding which source sentences are clinically or scientifically salient. These results support vote averaging as a stable scoring procedure across diverse domains, from short radiology reports to long discharge summaries.

Importantly, this analysis measures self-consistency across stochastic replicates of the same judge model, not clinical correctness. Oracle--human agreement in Appendix~\ref{app:oracle-validation} separately evaluates alignment with clinician annotations.

\section{Sentence-Level Flagging vs.\ Document-Level Abstention}
\label{app:sent-vs-doc}

Conformal abstention methods control risk by withholding entire outputs \citep{abbasi-yadkori-abstention-2024,quach-clm-2024}. We instantiate document-level abstention for hallucination control with two aggregates of sentence support, $a(X)=\min_i\hat p_{\mathrm{sup}}(v_i)$ and $a(X)=\mathrm{mean}_i\,\hat p_{\mathrm{sup}}(v_i)$. 

A summary is released if $a(X)>t$ and rejected otherwise, with loss
% \[
% L_{\mathrm{doc}}(t,X)=\1\{\text{summary contains an unsupported sentence}\ \wedge\ a(X)>t\}.
% \]

\[
L_{\mathrm{doc}}(t,X)=\1\left\{
\begin{gathered}
\text{summary contains an}\\
\text{unsupported sentence}\ \wedge\ a(X)>t
\end{gathered}
\right\}.
\]

This loss is bounded and non-increasing in $t$ for either aggregate, and we calibrate $t$ with the same scalar CRC rule as \textsc{CARE}: the smallest $t$ with $\frac{n}{n+1}\widehat R(t)+\frac{1}{n+1}\le\alpha$. Both baselines therefore have valid CRC guarantees. A violation above $\alpha$ on a single test split (e.g., 18.7\% for mean support on MIMIC-BHC) is consistent with finite-sample variation under a valid procedure, so validity must be assessed over repeated splits. Table~\ref{tab:sent-vs-doc} reports both baselines over the same 100 resplits as \textsc{CARE}.

\begin{table}[!htbp]
\centering
\scriptsize
\setlength{\tabcolsep}{3pt}
\caption{Sentence-level flagging vs.\ document-level abstention for hallucination control at $\alpha=0.15$, 100 resplits. Each abstention cell reports mean violation \% / \% of summaries rejected. \textsc{CARE} cells report mean violation \% / \% of summary sentences flagged. All three procedures are valid CRC procedures. $^\ast$Within Monte Carlo error of $\alpha$; at 2{,}000 resplits, mean-support violation is 14.6--15.0\% on all datasets.}
\label{tab:sent-vs-doc}
\resizebox{0.48\textwidth}{!}{%
\begin{tabular}{l ccc}
\toprule
Dataset & Doc, $\min\hat p_{\mathrm{sup}}$ & Doc, $\mathrm{mean}\,\hat p_{\mathrm{sup}}$ & \textsc{CARE} (sentence) \\
\midrule
ACI-Bench & 5.8 / 83.6 & 15.9$^\ast$ / 59.3 & 12.2 / 13.7 \\
MIMIC-BHC & 3.0 / 78.8 & 15.2$^\ast$ / 56.8 & 12.5 / 13.5 \\
MIMIC-CXR & 2.5 / 68.1 & 14.9 / 28.7 & 5.7 / 24.4 \\
Priv-DS & 12.2 / 81.2 & 14.7 / 78.5 & 12.8 / 22.6 \\
SumPubMed & 11.3 / 28.5 & 14.6 / 27.5 & 6.8 / 7.7 \\
\bottomrule
\end{tabular}}
\end{table}

\noindent
The comparison is therefore about efficiency and actionability rather than validity. To control document-level hallucination risk, min- and mean-support abstention reject 28--84\% and 28--79\% of entire summaries. These rejections include 6--73\% and 10--44\% of summaries that contain no hallucination. \textsc{CARE} achieves the same type of guarantee by flagging 8--24\% of summary sentences while retaining the full draft, so clinicians review specific claims rather than discarding or rewriting whole summaries.

\section{Summarizer Transfer}
\label{app:summarizer-transfer}

To test whether \textsc{CARE} transfers across summarization models, we replace the default Llama-3.3-70B-Instruct summarizer with Gemini 2.5 Pro, GPT-5, and GPT-5-mini. For each summarizer, we rerun the full pipeline on ACI-Bench and MIMIC-BHC, including oracle labeling, judge scoring, risk-controlled calibration, and evaluation. The oracle labeler (GPT-5) and deployment-time judge (GPT-5-mini) remain fixed. Llama-3.3-70B and Gemini 2.5 Pro do not overlap with either evaluation model. The GPT-5 and GPT-5-mini summarizers do share a model with the oracle or judge, so these rows test transfer under possible self-evaluation bias. All results are averaged over 100 random 70/30 calibration/test resplits at $\alpha=0.15$.

\begin{table}[!htbp]
\centering
\small
\caption{Summarizer transfer at $\alpha=0.15$, averaged over 100 random 70/30 calibration/test resplits. \textsc{CARE} is recalibrated separately for each summarizer. Brackets denote a descriptive 95\% bootstrap interval over resplit-level estimates. \emph{Fl\%}: percentage of summary sentences flagged for hallucination or source sentences surfaced for omission. $^\ddagger$Shares a model with the oracle (GPT-5) or judge (GPT-5-mini).}
\label{tab:summarizer-transfer}
\resizebox{0.7\textwidth}{!}{
\begin{tabular}{ll ccc ccc}
\toprule
& & \multicolumn{3}{c}{\textbf{Hallucination}} & \multicolumn{3}{c}{\textbf{Omission}} \\
\cmidrule(lr){3-5} \cmidrule(lr){6-8}
Dataset & Summarizer & Viol.\% & Fl\% & Recall\% & Viol.\% & Fl\% & Recall\% \\
\midrule
ACI-Bench & Llama-3.3-70B & 12.2{\tiny [10.9,13.4]} & 13.7 & 87.5 & 5.7{\tiny [5.0,6.5]} & 29.3 & 88.7 \\
\rowcolor{gray!6}
& Gemini 2.5 Pro & 13.4{\tiny [12.2,14.6]} & 37.8 & 95.1 & 4.9{\tiny [4.2,5.6]} & 45.8 & 89.4 \\
& GPT-5$^\ddagger$ & 13.5{\tiny [12.1,14.8]} & 62.0 & 99.0 & 6.4{\tiny [5.7,7.1]} & 29.8 & 86.0 \\
\rowcolor{gray!6}
& GPT-5-mini$^\ddagger$ & 14.5{\tiny [13.3,15.7]} & 48.3 & 97.3 & 5.6{\tiny [4.8,6.5]} & 44.5 & 89.1 \\
\midrule
MIMIC-BHC & Llama-3.3-70B & 12.5{\tiny [12.1,13.0]} & 13.5 & 85.7 & 10.4{\tiny [10.0,10.7]} & 62.0 & 89.5 \\
\rowcolor{gray!6}
& Gemini 2.5 Pro & 12.3{\tiny [11.9,12.8]} & 36.8 & 94.5 & 9.6{\tiny [9.4,9.8]} & 67.9 & 90.9 \\
& GPT-5$^\ddagger$ & 12.7{\tiny [12.2,13.2]} & 44.8 & 95.2 & 8.7{\tiny [8.5,8.8]} & 57.6 & 90.7 \\
\rowcolor{gray!6}
& GPT-5-mini$^\ddagger$ & 11.1{\tiny [10.5,11.8]} & 40.9 & 91.5 & 9.7{\tiny [9.3,10.2]} & 63.6 & 89.9 \\
\bottomrule
\end{tabular}}
\end{table}

Table~\ref{tab:summarizer-transfer} shows that, after recalibration, \textsc{CARE} keeps mean violation at or below the target risk level for all four summarizers and both controllers. Workload depends strongly on the summarizer. Llama-3.3-70B requires the fewest hallucination flags (13.5--13.7\% of summary sentences). Gemini 2.5 Pro, GPT-5, and GPT-5-mini require 37--62\%, which suggests that their summaries contain more sentences the judge scores as only partially supported. Omission workload varies less, except that Gemini 2.5 Pro and GPT-5-mini surface more source content on ACI-Bench. These results show that the same calibration procedure applies across summarizers after representative recalibration. They should not be read as plug-and-play threshold transfer.

\section{Distribution Shift Robustness}
\label{app:distribution-shift}

\textsc{CARE}'s formal guarantees assume matched calibration and deployment distributions. We therefore evaluate two empirical stress tests that deliberately violate this assumption: cross-dataset threshold transfer and within-dataset length shift. Table~\ref{tab:distribution-shift} reports the cross-dataset and within-dataset length-shift results. The formal CRC and LTT-FST guarantees do not apply in these shifted settings.

\begin{table}[!t]
\centering
\small
\caption{Distribution shift experiments at $\alpha=0.15$, averaged over 100 random resplits. (a) Cross-dataset threshold transfer for omission control, rows indicate calibration dataset and columns indicate test dataset. Shaded diagonal entries are in-distribution. (b) Within-dataset length shift, calibrating on short documents and testing on long documents, or vice versa.}
\label{tab:distribution-shift}
\vspace{4pt}
{\centering\textbf{(a) Cross-dataset transfer (omission frac.\ violation \%)}\par}
\vspace{2pt}
\resizebox{0.48\textwidth}{!}{%
\begin{tabular}{l ccccc}
\toprule
Cal $\backslash$ Test & ACI & BHC & CXR & Priv-DS & PubMed \\
\midrule
  ACI & \cellcolor{gray!20}\textbf{5.2} & 5.2 & 0.4 & 4.6 & 5.5 \\
  BHC & 7.9 & \cellcolor{gray!20}\textbf{10.3} & 0.4 & 6.5 & 6.7 \\
  CXR & 10.4 & 5.9 & \cellcolor{gray!20}\textbf{2.1} & 5.7 & 11.1 \\
  Priv-DS & 9.5 & 15.5$^\dagger$ & 0.6 & \cellcolor{gray!20}\textbf{9.1} & 7.9 \\
  PubMed & 8.3 & 11.6 & 0.4 & 7.1 & \cellcolor{gray!20}\textbf{7.4} \\
\bottomrule
\end{tabular}%
}
\vspace{8pt}

\textbf{(b) Within-dataset length shift (violation \%)}
\vspace{2pt}
\noindent\makebox[\linewidth][c]{%
\resizebox{0.6\textwidth}{!}{%
\begin{tabular}{l ccc ccc}
\toprule
 & \multicolumn{3}{c}{Hallucination} & \multicolumn{3}{c}{Omission (frac.)} \\
\cmidrule(lr){2-4} \cmidrule(lr){5-7}
Dataset & Short$\to$Long & Long$\to$Short & Random & Short$\to$Long & Long$\to$Short & Random \\
\midrule
    ACI & 11.4 & 12.2 & 12.4 & 3.7 & 1.5 & 5.4 \\
  BHC & 14.6 & 10.5 & 12.7 & 6.7 & 7.4 & 10.1 \\
  CXR & 5.2 & 6.4 & 5.9 & 2.1 & 1.7 & 1.9 \\
  Priv-DS & 17.5$^\dagger$ & 8.5 & 12.7 & 7.6 & 9.0 & 9.3 \\
  PubMed & 8.0 & 6.9 & 6.9 & 8.2 & 6.6 & 7.5 \\
\bottomrule
\end{tabular}%
}%
}
\vspace{2pt}
\par\small $^\dagger$Exceeds $\alpha=0.15$. Shaded = in-distribution (diagonal).
\end{table}

\paragraph{Cross-dataset transfer}
Cross-dataset transfer of LTT-FST omission thresholds is less fragile than its margin below $\alpha$ might suggest. Off-diagonal violation is at or below $\alpha=0.15$ in 19 of 20 transfers. The one exception is Priv-DS$\to$MIMIC-BHC at 15.5\%. Because LTT-FST in-distribution operating points sit well below $\alpha$, this margin absorbs much of the shift. Transferred thresholds are not efficient, however: they were certified for a different workload--risk tradeoff, and the guarantee does not apply off-distribution. \textsc{CARE} should therefore be recalibrated when the deployment domain differs from the calibration set.

\paragraph{Within-dataset length shift}
For hallucination control, calibrating on short documents and testing on long documents causes a violation on Priv-DS (17.5\%), while calibrating on long documents and testing on short documents remains below $\alpha$ for all datasets. This pattern is expected: short-document calibration can select thresholds that are insufficient for longer documents with more opportunities for error. Omission control remains below $\alpha$ under both length shifts on all datasets. Because the formal guarantee does not apply under shift, representative recalibration remains necessary when deployment differs in document length or complexity.

\section{Computational Costs}
\label{app:cost}

Table~\ref{tab:cost} reports estimated API costs and token usage for the \textsc{CARE} pipeline, normalized to 100 documents for calibration and to one document for deployment. Calibration is an offline, one-time cost that includes summarization, oracle labeling, and judge scoring. Deployment excludes oracle labeling and requires only summarization, judge scoring, and thresholding. Estimated calibration cost ranges from \$0.58 for MIMIC-CXR to \$113.89 for Priv-DS per 100 documents. Deployment cost ranges from \$0.003 to \$0.867 per document, depending primarily on source length. Thresholding itself adds negligible computational overhead once sentence scores are computed. For long clinical documents, deployment-time latency is likely best handled through asynchronous or pre-signature review workflows, where summaries and safety flags can be generated before clinician review.

\begin{table*}[!t]
\centering
\small
\caption{Estimated API costs and token usage for \textsc{CARE}. Calibration costs are reported for 100 documents, deployment costs are per document. \textbf{Fact}, \textbf{Imp}, and \textbf{Cov} denote factuality, importance, and coverage scoring. $^\ddagger$ACI-Bench has fewer than 100 calibration documents, so totals are mean-padded to 100 for comparability.}
\label{tab:cost}
\vspace{0.5em}
\textbf{(a)} API costs (USD).
\vspace{0.3em}

\begin{tabular}{l rrrrr rrrrr}
\toprule
& \multicolumn{5}{c}{\textbf{Calibration (100 docs)}} & \multicolumn{5}{c}{\textbf{Deployment (per doc)}} \\
\cmidrule(lr){2-6} \cmidrule(lr){7-11}
\textbf{Dataset} & Summ. & Fact & Imp & Cov & \textbf{Total} & Summ. & Fact & Imp & Cov & \textbf{Total} \\
\midrule
ACI$^\ddagger$ & 0.15 & 1.53 & 2.13 & 1.78 & 5.59 & 0.001 & 0.008 & 0.014 & 0.009 & 0.032 \\
BHC & 0.26 & 1.63 & 6.90 & 3.57 & 12.35 & 0.003 & 0.008 & 0.047 & 0.018 & 0.076 \\
CXR & 0.02 & 0.18 & 0.19 & 0.19 & 0.58 & <0.001 & 0.001 & 0.001 & 0.001 & 0.003 \\
Priv-DS & 1.29 & 13.23 & 85.56 & 13.81 & 113.89 & 0.013 & 0.066 & 0.719 & 0.069 & 0.867 \\
PubMed & 0.45 & 1.91 & 14.76 & 5.67 & 22.80 & 0.005 & 0.010 & 0.121 & 0.028 & 0.163 \\
\bottomrule
\end{tabular}
\par\vspace{1.5em}
\textbf{(b)} Token usage.
Calibration: millions of tokens for 100 documents. Deployment: thousands of tokens per document.
\vspace{0.3em}

\begin{tabular}{l rrrrr rrrrr}
\toprule
& \multicolumn{5}{c}{\textbf{Calibration (M tokens, 100 docs)}} & \multicolumn{5}{c}{\textbf{Deployment (K tokens, per doc)}} \\
\cmidrule(lr){2-6} \cmidrule(lr){7-11}
\textbf{Dataset} & Summ. & Fact & Imp & Cov & \textbf{Total} & Summ. & Fact & Imp & Cov & \textbf{Total} \\
\midrule
ACI$^\ddagger$ & 0.21 & 0.96 & 1.19 & 0.92 & 3.28 & 2.1 & 4.8 & 8.7 & 4.6 & 20.2 \\
BHC & 0.37 & 1.14 & 4.38 & 1.72 & 7.61 & 3.7 & 5.7 & 31.9 & 8.6 & 49.9 \\
CXR & 0.03 & 0.10 & 0.07 & 0.08 & 0.28 & 0.3 & 0.5 & 0.4 & 0.4 & 1.6 \\
Priv-DS & 1.82 & 10.24 & 64.89 & 7.48 & 84.43 & 18.2 & 51.2 & 557.0 & 37.4 & 663.8 \\
PubMed & 0.64 & 1.40 & 10.10 & 2.82 & 14.96 & 6.4 & 7.0 & 87.9 & 14.1 & 115.4 \\
\bottomrule
\end{tabular}
\par\vspace{1.5em}
\textbf{(c)} Deployment-time judge API cost (USD per document, local-compute cost excluded).
External scorers (NLI cross-encoder, BERT embedding cosine, AlignScore) and Qwen3.5-4B run as local GPU inference and incur no per-token API cost. Gemini 2.5 Pro was run on the two datasets where its outputs are available (App.~\ref{app:scorer-agnostic}), ``--'' marks datasets where the experiment was not run.
\vspace{0.3em}

\begin{tabular}{l c c c c}
\toprule
\textbf{Dataset} & GPT-5-mini & Gemini 2.5 Pro & Qwen3.5-4B$^{\dagger}$ & External$^{\dagger}$ \\
\midrule
ACI$^\ddagger$ & 0.031 & 0.154 & 0.000 & 0.000 \\
BHC & 0.073 & 0.365 & 0.000 & 0.000 \\
CXR & 0.003 & -- & -- & 0.000 \\
Priv-DS & 0.854 & -- & -- & 0.000 \\
PubMed & 0.158 & -- & -- & 0.000 \\
\bottomrule
\end{tabular}
\par\vspace{0.2em}
\footnotesize $^{\dagger}$Run with local GPU inference. No per-token API charge is included, but local compute cost is not zero.\\

\end{table*}

\section{Clinician Study}
\label{app:clinician-study}

\paragraph{Design.} We conducted a preliminary randomized within-reader study to evaluate whether \textsc{CARE} increases review-time coverage of omitted content. Three clinicians (C1--C3) each reviewed 25 de-identified documents drawn from ACI-Bench and MIMIC-BHC, for 75 reviews in total. Each document was randomly assigned to a \textsc{CARE}-assisted condition, in which omission and hallucination flags were displayed, or to an unassisted control condition with no annotations. The arms contained 37 reviews (16 ACI-Bench, 21 MIMIC-BHC) and 38 reviews (14 ACI-Bench, 24 MIMIC-BHC), respectively. Clinicians read the source and the LLM-generated summary, highlighted source sentences they considered important but missing, and edited the summary. We recorded review time, highlighted sentences, and edited summaries. The flags were generated at the time of the study (March--April 2026) by an earlier version of the pipeline, from an earlier version of the judge scores. The exact flagged and surfaced sentences shown to each clinician are recorded in the study logs, and all study metrics use these recorded sets. Recomputing the surfaced sets from the current judge scores does not reproduce the recorded ones, so the study evaluates the review interface at its own historical operating point rather than at the thresholds reported in this paper. The three clinicians are co-authors who served as study annotators; no external participants were recruited.

\paragraph{Endpoint decomposition.} Let $T$ denote oracle-labeled true omissions, $Y$ the clinician's highlights, and $S$ \textsc{CARE}'s surfaced sentences. The primary endpoint, assisted review-time coverage, is $\lvert T\cap(Y\cup S)\rvert/\lvert T\rvert$. In the control arm, $S=\varnothing$. Table~\ref{tab:clinician-study} reports this endpoint per clinician, and Table~\ref{tab:clinician-decomp} decomposes it. $p$-values are from one-sided permutation tests that permute arm assignment within clinician, and intervals are bootstrap intervals over reviews.

All three clinicians had higher assisted coverage with \textsc{CARE}, and all three also highlighted more true omissions themselves. The aggregate gain in clinicians' own detection (+9.9 points) was not statistically significant. With only three readers, we treat these comparisons as descriptive rather than as population-level inference over clinicians.

\paragraph{Responses to flags and edit behavior.} Clinicians independently highlighted 52.2\% of \textsc{CARE}'s omission flags. Measured against oracle labels, these flags were at least as precise as clinicians' own highlights (17.7\% vs.\ 15.2\%), and there were fewer of them per review. Absolute precision is low for both. Clinicians removed or rewrote 56\% of hallucinated summary sentences with \textsc{CARE} vs.\ 47\% without it. However, they edited flagged and unflagged sentences at similar rates (51\% vs.\ 58\%), so this difference cannot be attributed to the flags. An automated GPT-5 re-labeling of 70 edited summaries showed that editing reduced the omission rate by about 12 percentage points in both arms, with no arm difference. The study did not log explicit accept or reject actions per flag, and the final edited summaries were not scored by clinicians.

\paragraph{Post-session survey.} After each of their two sessions, clinicians rated the annotations on a five-point scale (six responses in total). Mean ratings were 3.7 for the helpfulness of hallucination flags, 3.5 for the helpfulness of omission highlights, 3.8 for trust in the annotations, and 3.3 for their effect on confidence. Free-text comments were mixed. One clinician found the omission highlights more reliable and another found the hallucination flags more helpful. A third noted that they rewrote most summaries from scratch and therefore relied on the annotations less, which suggests that annotation value depends on summary quality and the reviewer's editing style.

\paragraph{Scope.} The study includes only three clinicians and uses benchmark review tasks rather than real clinical deployment, so we interpret these findings as preliminary workflow evidence. Larger prospective studies should measure final documentation quality, flag-level feedback, trust, alert burden, and review behavior.

\begin{table}[!htbp]
\centering
\small
\caption{Assisted review-time omission coverage by clinician: the percentage of oracle-labeled true omissions identified during review, either through clinician highlighting or, in the \textsc{CARE}-assisted condition, through a \textsc{CARE} surfacing flag. Each clinician reviewed 25 documents.}
\label{tab:clinician-study}
\begin{tabular}{l c rr r}
\toprule
 & & \multicolumn{3}{c}{Assisted coverage (\%)} \\
\cmidrule(lr){3-5}
 & $n$ & Control & \textsc{CARE} & $\Delta$\,(pp) \\
\midrule
C1 & 25 & 70 & 84 & $+14$ \\
C2 & 25 & 63 & 84 & $+22$ \\
C3 & 25 & 20 & 68 & $+48$ \\
\midrule
\textbf{All clinicians} & \textbf{75} & \textbf{50.4} & \textbf{79.0} & $\mathbf{+28.6}$ \\
\bottomrule
\end{tabular}
\end{table}

\begin{table}[!htbp]
\centering
\scriptsize
\setlength{\tabcolsep}{3pt}
\caption{Decomposition of the clinician-study endpoint (means over reviews; 95\% bootstrap intervals).}
\label{tab:clinician-decomp}
\resizebox{0.55\textwidth}{!}{%
\begin{tabular}{l c c}
\toprule
Metric & Control & \textsc{CARE} \\
\midrule
Assisted coverage $\lvert T\cap(Y\cup S)\rvert/\lvert T\rvert$ & 50.4\% [41, 60] & 79.0\% [72, 85], $p<0.001$ \\
Clinicians' own highlights $\lvert T\cap Y\rvert/\lvert T\rvert$ & 50.4\% [41, 60] & 60.3\% [51, 70], $p=0.08$ \\
\quad per clinician (C1 / C2 / C3) & 70 / 63 / 20\% & 71 / 76 / 33\% \\
\textsc{CARE} flags alone $\lvert T\cap S\rvert/\lvert T\rvert$ & -- & 56.7\% [46, 67] \\
Found by clinicians beyond flags & -- & 22.3\% [15, 30] \\
Flags also highlighted by clinician & -- & 52.2\% [44, 60] \\
Precision vs.\ oracle: flags / own highlights & -- / 15.2\% & 17.7\% / 15.2\% \\
Items per review: highlights / flags & 33.2 / -- & 33.6 / 25.1 \\
Seconds per source sentence & 8.40 [7.3, 9.6] & 7.95 [7.2, 8.8] \\
Minutes per review & 11.4 [9.9, 13.1] & 11.3 [10.1, 12.6] \\
\bottomrule
\end{tabular}}
\end{table}

\section{Worked Example (ACI-Bench)}
\label{app:example}

Tables~\ref{tab:example-summary} and~\ref{tab:example-source} show the complete \textsc{CARE} output for one ACI-Bench document from the canonical test split: every generated summary sentence with its support score and hallucination decision, and every source dialogue turn with its importance and non-coverage scores and omission decision, alongside the oracle labels. The example is generated by \texttt{paper/scripts/example\_aci.py} at the thresholds reported in Table~\ref{tab:thresholds}. Both unsupported summary sentences are flagged. Four supported plan sentences are also flagged because the judge gave them low support scores; they are false positives that a clinician would review and keep. Both true omissions are surfaced: the doctor's statement that there are no vision problems, and the reassurance about the cataract surgery and skin-cancer removal. Four further turns are surfaced that the oracle considers covered. The judge was unsure they were covered ($\hat p_{\mathrm{ncov}}\ge0.10$), and the low deployed coverage threshold ($\gamma^*=0.05$) passes such turns. A clinician reviewing this summary would read 6 of 33 summary sentences and 6 of 38 dialogue turns rather than the full encounter.

\begin{table*}[!p]
\centering
\scriptsize
\setlength{\tabcolsep}{3pt}
\caption{Worked example, part 1: generated summary with hallucination flags. Document 73 (canonical test split). Thresholds from the canonical calibration split: $\lambda^*=0.70$, $(\tau^*,\gamma^*)=(0.15,0.05)$. A summary sentence is flagged when $\hat p_{\mathrm{sup}}\le\lambda^*$. The oracle labels 2 sentence(s) unsupported (shaded); \textsc{CARE} flags 6 sentences, 2 of which are the unsupported ones (TP); the rest are false positives (FP) routed to review.}
\label{tab:example-summary}
\begin{tabular}{r p{11.4cm} c c l}
\toprule
\# & Summary sentence & $\hat p_{\mathrm{sup}}$ & Oracle & \textsc{CARE} \\
\midrule
1 & Patient Name: Louis. & 1.00 & Supp. &  \\
2 & Age: 58. & 1.00 & Supp. &  \\
3 & Sex: Male. & 1.00 & Supp. &  \\
4 & Chief Complaint: Shortness of breath and dizziness after playing tennis in hot and humid weather. & 1.00 & Supp. &  \\
5 & History of Present Illness: The patient presented to the emergency room with shortness of breath and was found to have high blood pressure. & 1.00 & Supp. &  \\
6 & He has a history of congestive heart failure and has been taking his medications as prescribed. & 1.00 & Supp. &  \\
7 & He reported eating salty foods occasionally, which may have contributed to the exacerbation of his heart failure. & 1.00 & Supp. &  \\
8 & * Congestive heart failure. & 1.00 & Supp. &  \\
9 & * Cataract removal (3 months ago). & 1.00 & Supp. &  \\
10 & * Skin cancer removal (5 months ago). & 1.00 & Supp. &  \\
11 & * Current medications for heart failure (not specified). & 1.00 & Supp. &  \\
12 & * New medications: & 1.00 & Supp. &  \\
13 & + Lasix 80mg once daily. & 1.00 & Supp. &  \\
14 & + Carvedilol 25mg twice daily. & 1.00 & Supp. &  \\
15 & * Blood pressure: elevated. & 1.00 & Supp. &  \\
16 & * Neck: supple. & 1.00 & Supp. &  \\
17 & * Heart: regular rate and rhythm, no murmur. & 1.00 & Supp. &  \\
18 & * Lungs: fine crackles bilaterally. & 1.00 & Supp. &  \\
19 & * Lower extremities: trace edema. & 1.00 & Supp. &  \\
20 & * ECG: stable, showing left ventricular hypertrophy. & 1.00 & Supp. &  \\
21 & * Echocardiogram: low pumping function, but stable. & 1.00 & Supp. &  \\
22 & * Skin biopsy: no evidence of residual skin cancer. & 1.00 & Supp. &  \\
23 & * Exacerbation of congestive heart failure due to dietary indiscretion (eating salty foods). & 1.00 & Supp. &  \\
24 & * High blood pressure. & 1.00 & Supp. &  \\
25 & * Add Lasix 80mg once daily to help with fluid retention. & 1.00 & Supp. &  \\
26 & * Add Carvedilol 25mg twice daily to help with blood pressure control. & 0.50 & Supp. & \textcolor{redflag}{\textbf{flag}} (FP) \\
27 & * Continue to avoid salty foods and follow a low-sodium diet. & 0.50 & Supp. & \textcolor{redflag}{\textbf{flag}} (FP) \\
28 & \cellcolor{lightred}* Monitor blood pressure and report any changes to the physician. & 0.10 & Unsupp. & \textcolor{redflag}{\textbf{flag}} \textbf{(TP)} \\
29 & * Schedule a follow-up appointment to monitor the patient's condition and adjust medications as needed. & 0.00 & Supp. & \textcolor{redflag}{\textbf{flag}} (FP) \\
30 & * Lasix 80mg once daily. & 1.00 & Supp. &  \\
31 & * Carvedilol 25mg twice daily. & 1.00 & Supp. &  \\
32 & \cellcolor{lightred}* The nurse will come in to review the patient's medications and provide instructions on how to take them. & 0.50 & Unsupp. & \textcolor{redflag}{\textbf{flag}} \textbf{(TP)} \\
33 & * The patient will be monitored for any changes in his condition and report any concerns to the physician. & 0.10 & Supp. & \textcolor{redflag}{\textbf{flag}} (FP) \\
\bottomrule
\end{tabular}
\end{table*}

\begin{table*}[p]
\centering
\scriptsize
\renewcommand{\arraystretch}{0.85}
\setlength{\tabcolsep}{3pt}
\caption{Worked example, part 2: source dialogue with omission surfacing for the same document. A source turn is surfaced when $\hat p_{\mathrm{imp}}\ge\tau^*$ and $\hat p_{\mathrm{ncov}}\ge\gamma^*$. The oracle labels 2 turn(s) as true omissions (important and not covered; shaded); \textsc{CARE} surfaces 6 of 38 turns, including 2 of the 2 true omissions. Other surfaced turns are important content the judge was unsure the summary covered.}
\label{tab:example-source}
\begin{tabular}{r p{10.6cm} c c c l}
\toprule
\# & Source turn & $\hat p_{\mathrm{imp}}$ & $\hat p_{\mathrm{ncov}}$ & Oracle & \textsc{CARE} \\
\midrule
1 & [doctor] hi , louis . how are you ? & 0.00 & 1.00 & --/Omit &  \\
2 & [patient] hi . good to see you . & 0.00 & 1.00 & --/Omit &  \\
3 & [doctor] it's good to see you as well . are you ready to get started ? & 0.00 & 1.00 & --/Omit &  \\
4 & [patient] yes , i am . & 0.00 & 1.00 & --/Omit &  \\
5 & [doctor] louis is a 58-year-old male here for follow up from an emergency room visit . so , louis , what happened ? & 1.00 & 0.00 & Imp/Cov &  \\
6 & [patient] yeah . i was playing tennis on saturday . it was really , really hot that day , very humid . and about after about a half an hour i was very short of breath , i was struggling breathing . i thought i was having a heart attack , got really nervous . so , my wife took me to the er and , uh , everything checked out , but i was just very upset about it . & 1.00 & 0.50 & Imp/Cov & \textcolor{blueflag}{\textbf{surface}} \\
7 & [doctor] okay . all right . and how have you been feeling since that time ? & 0.00 & 0.80 & --/Omit &  \\
8 & [patient] uh , foof , probably , probably about six hours after we got home , i felt very light-head and very dizzy and then , sunday , i felt fine . i just thought it was worth checking up with you though . & 1.00 & 0.50 & --/Cov & \textcolor{blueflag}{\textbf{surface}} \\
9 & [doctor] okay . and have you been taking all of your meds for your heart failure ? & 0.00 & 0.00 & Imp/Cov &  \\
10 & [patient] i have . i have . i've been , uh , very diligent with it . and , uh , i'm in touch with the doctor and so far , so good , other than this episode on saturday . & 0.90 & 0.10 & --/Cov & \textcolor{blueflag}{\textbf{surface}} \\
11 & [doctor] okay . and , and you're watching your diet , you're avoiding salt . have you had anything salty ? & 0.00 & 0.00 & --/Cov &  \\
12 & [patient] i cheat every now and then . you know , i try and stay away from the junk food and the salty foods . but , for the most part , i've been doing a good job of that . & 0.90 & 0.00 & Imp/Cov &  \\
13 & [doctor] okay . all right . um , and i know that they removed a cataract from your eye- & 0.80 & 0.00 & Imp/Cov &  \\
14 & [doctor] . a couple of , like couple months ago . that's been fine ? & 0.80 & 0.00 & --/Cov &  \\
15 & [patient] that was three months ago , thursday , and everything's been fine ever since . & 0.80 & 0.00 & Imp/Cov &  \\
16 & \cellcolor{purplebg}[doctor] okay . so , no vision problems . & 0.50 & 1.00 & Imp/Omit & \textcolor{blueflag}{\textbf{surface}} \textbf{(TP)} \\
17 & [patient] no . & 0.00 & 1.00 & --/Omit &  \\
18 & [doctor] okay . and you had a skin cancer removed about five months ago as well . you've had a lot going on . & 0.60 & 0.00 & Imp/Cov &  \\
19 & [patient] yeah . it's been a really busy year . an- and again , so far , so good . that healed up nicely , no problems ever since . & 0.50 & 0.30 & Imp/Cov & \textcolor{blueflag}{\textbf{surface}} \\
20 & [doctor] okay . all right . um , so , why do n't we go ahead and we'll do a quick physical- & 0.00 & 0.20 & --/Cov &  \\
21 & [doctor] . exam . hey , dragon , show me the blood pressure . so , here , your blood pressure is a little high . & 1.00 & 0.00 & --/Cov &  \\
22 & [doctor] um , so , you know , i did see a report in the emergency room that your blood pressure was high there as well . & 0.80 & 0.00 & --/Cov &  \\
23 & [doctor] so , we'll have to just kind of talk about that . uh , but let's go ahead and we'll examine you . & 0.10 & 0.20 & --/Cov &  \\
24 & [patient] sure . & 0.00 & 1.00 & --/Omit &  \\
25 & [doctor] okay ? & 0.00 & 1.00 & --/Omit &  \\
26 & [doctor] okay . so , you know , looking at you , your neck is very supple . i do n't appreciate any fibular venous distention . your heart is a regular rate and rhythm , no murmur . your lungs have some fine crackles in them , bilaterally . and you have trace lower extremity edema in both legs . so , what that means , essentially , is that you may have some extra fluid on board , um , from eating salty foods- & 1.00 & 0.00 & Imp/Cov &  \\
27 & [doctor] . along with this history of your congestive heart failure . but , let's go ahead and look at some of your results . hey , dragon , show me the ecg . so , this is , uh , a s- a stable ecg for you . this basically shows that you have some left ventricular hypertrophy which caused your congestive heart failure . um , let's go ahead and review your echocardiogram . hey , dragon , show me the echocardiogram . so , in reviewing the results of your echocardiogram , it shows that your pumping function of your heart is a little low , uh , but it's stable . and , you know , i think that we know this and we have you on the appropriate- & 1.00 & 0.00 & Imp/Cov &  \\
28 & [doctor] medication therapy . and then , i just wan na be reminded about , um , the results of your skin biopsy . hey , dragon , show me the skin biopsy results . okay . and in reviewing the pathology report for your skin cancer- & 0.50 & 0.00 & --/Cov &  \\
29 & [doctor] . you know , it looks like they got all of that and everything's fine . & 0.70 & 0.00 & Imp/Cov &  \\
30 & [patient] yep . & 0.00 & 1.00 & --/Omit &  \\
31 & [doctor] so , you know , my impression of you at this time , for the shortness of breath that you had in the emergency department , i think it was an exacerbation of your heart failure . and you probably had some , what we call , dietary indiscretion , you ate some salty food which made you retain some fluid . & 1.00 & 0.00 & Imp/Cov &  \\
32 & [doctor] so , for that , i'm going to prescribe you , you know , an extra dose of lasix 80 milligrams , once a day . and , um , we're going to , uh , put you on some carvedilol 25 milligrams , twice a day . okay ? & 1.00 & 0.00 & Imp/Cov &  \\
33 & [patient] okay . perfect . & 0.00 & 1.00 & --/Omit &  \\
34 & \cellcolor{purplebg}[doctor] um , and i think from a , a cataract surgery standpoint and your skin cancer removal , everything seems to be fine and you're doing well , so i do n't think we need to adjust any of those medications . & 1.00 & 0.50 & Imp/Omit & \textcolor{blueflag}{\textbf{surface}} \textbf{(TP)} \\
35 & [patient] good to hear . & 0.00 & 1.00 & --/Omit &  \\
36 & [doctor] okay ? hey , dragon , order lasix 80 milligrams , once a day . hey , dragon , order carvedilol 25 milligrams , twice a day . okay . and the nurse will come in and she'll see you soon . okay ? & 1.00 & 0.00 & Imp/Cov &  \\
37 & [patient] great . & 0.00 & 1.00 & --/Omit &  \\
38 & [doctor] hey , dragon , finalize the note . & 0.00 & 1.00 & --/Omit &  \\
\bottomrule
\end{tabular}
\end{table*}

\end{document}